\documentclass[final,3p,times]{elsarticle}

\usepackage{graphicx}
\usepackage{amssymb}

\usepackage{url}
\usepackage{makecell}
\usepackage{bm}
\usepackage{mathtools}

\usepackage{caption}
\usepackage{subcaption}
\usepackage{tabularx}
\usepackage{booktabs}

\usepackage{diagbox}

\usepackage[utf8]{inputenc}
\usepackage{float}
\usepackage{graphicx}
\usepackage{multirow} 
\usepackage{subcaption}
\usepackage{breakcites} 

\usepackage{xcolor}

\journal{Elsevier}

\begin{document}

\begin{frontmatter}

\title{Prediction of rare feature combinations in population synthesis: Application of deep generative modelling}

\author{Sergio Garrido}
\ead{shgm@dtu.dk}
\author{Stanislav S. Borysov}
\author{Francisco C. Pereira}
\author{Jeppe Rich}
\address{Department of Technology, Management and Economics, Technical University of Denmark, DTU, 2800 Kgs. Lyngby, Denmark}

\begin{abstract}
In population synthesis applications, when considering populations with many attributes, a fundamental problem is the estimation of rare combinations of feature attributes. Unsurprisingly, it is notably more difficult to reliably represent the sparser regions of such multivariate distributions and in particular combinations of attributes which are absent from the original sample. In the literature this is commonly known as sampling zeros for which no systematic solution has been proposed so far. In this paper, two machine learning algorithms, from the family of deep generative models, are proposed for the problem of population synthesis and with particular attention to the problem of sampling zeros. Specifically, we introduce the Wasserstein Generative Adversarial Network (WGAN) and the Variational Autoencoder (VAE), and adapt these algorithms for a large-scale population synthesis application. The models are implemented on a Danish travel survey with a feature-space of more than 60 variables. The models are validated in a cross-validation scheme and a set of new metrics for the evaluation of the sampling-zero problem is proposed. Results show how these models are able to recover sampling zeros while keeping the estimation of truly impossible combinations, the \emph{structural zeros}, at a comparatively low level. Particularly, for a low dimensional experiment, the VAE, the marginal sampler and the fully random sampler generate 5\%, 21\% and 26\%, respectively, more structural zeros per sampling zero generated by the WGAN, while for a high dimensional case, these figures escalate to 44\%, 2217\% and 170440\%, respectively. This research directly supports the development of agent-based systems and in particular cases where detailed socio-economic or geographical representations are required.

\end{abstract}

\parskip=8pt   
\parindent=0in 


\end{frontmatter}



\section{Introduction}
\label{sec:intro}

Population synthesis is concerned with the creation of realistic yet artificial populations of agents based on a real sample of such agents \citep{muller_2011_populationSynthesisSOTA, harland_2012_microsimulationReview}. The aim is to create populations such that the generated agents have statistical properties similar to the agents from the real population and possibly such that, at later stages, the artificially generated population can be aligned with variables that represent future targets.  

A simple solution to the problem of recovering the statistical properties of a sample is to apply random sampling with replacement. Although it will reproduce the statistical properties of the sample, it is undesirable for two main reasons. First, due to data protection laws, it is generally not allowed to disclose information of agents in a way such that they might be personally identified. This is true, even if there are no individual identifiers, as detailed variable combinations of agents can make it possible to disclose specific individuals. Secondly, from a more technical perspective, making strict copies of the same individuals over and over again is not preferable either, as it is impossible to produce agents with a different combination of variables. If the samples on which the synthesis is based have a relatively small number of observations, but contain many attributes, then any re-sampling will suffer from the same type of sparsity as the sample. As population synthesis is widely used as a first step in Agent Based Models (ABMs) for which forecasting of demand is a common application \citep{balmer_2006_agent, bradley_2010_sacsim, rich_large_2018}, it is generally not preferable to use populations which are tied to specific (small) samples as the sparsity of the sample will be inherited in the ABM. In the extreme, this will imply that significant groups of agents that are completely missing from the sample, cannot be represented in the ABM. This problem is commonly referred to as the zero-cell or ``sampling zeros'' problem (\cite{cho_2014_syntheticPopulationTechniques}) and was first addressed by \cite{beckman_ipf_1996}, although from a numerical point of view. 

In this paper, we consider population synthesis from a model-based probabilistic perspective by exploring two novel machine learning models that enable the modeller to approximate the full joint distribution of high-dimensional data sets. The models represent a systematic way of circumventing the zero-cell problem by learning abstract representations \citep{goodfellow_dl_2016} from the available sample. In the paper we give specific attention to the zero-cell problem and introduce a set of performance indicators for the measurement of the zero-cell problem. The paper extends recent contributions by \cite{sun_bn_2015, sun_mixture_2018, saadi_hmm_2016, borysov_vae_2019}, for the first time using Generative Adversarial Networks (GAN) \citep{goodfellow_gan_2014, arjovsky_wgan_2017} and by comparing this approach with a Variational Auto-Encoder (VAE) \citep{kingma_vae_2014, rezende_dgm_2014} framework. 

\subsection{The role of population synthesis}

The increased use of Agent Based Models (ABMs) in a number of different research areas including economics,  ecology, environmental science and transport (\cite{donoghue_2014_survey}) has led to an upsurge in the application of population synthesis methods \citep{rasouli_2013_activity} as these methods enable modellers to synthesize and forecast populations of agents, which in subsequent stages can be used to simulate demand in these different contexts. Agent-based transport demand models that use synthetic populations have been presented in \cite{bradley_2010_sacsim, rich_danish_2016, miller_2015_implementation, fournier_2018_integration} and \cite{viegas_2018_modeling} to mention a few. Population synthesis is an important stage in the modelling process because it generates the basis for any demand investigation, namely a pool of agents with all the spatial and social features fundamental for the outcome of such investigation. The fact that population synthesis typically is prior to the agent-based model further underlines the importance of the population synthesis stage as errors accumulated at this stage will propagate in the model system and affect the demand forecasts \citep{krishnamurthy_uncertainty_2003}. 

\subsection{Methods used for population synthesis}

A classification of the different methodologies that have been applied to the problem of population synthesis is offered in  \cite{tanton_2014_review}. In this paper three main typologies are identified: i) sample expansion, ii) matrix fitting and iii) simulation-based approaches. While the current paper only considers methodologies belonging to the latter type, it is relevant to briefly describe how this branch of methods differs from the methodologies that have been applied historically. We can consider sample expansion and matrix fitting as belonging to a family of strictly deterministic methods and this is what differentiates these methods from the simulation methods. Deterministic methods consider the sample data as a ground truth distribution of features over a population and try to develop alternative populations by expanding the sample (by means of expansion factors) as in \cite{daly_1998_prototypical} or by fitting the sample to alternative marginal distributions typically by means of Iterative Proportional Fitting (IPF) \citep{deming_ipf_1940}. A second distinction is that these methods render outputs in the form of prototypical agents rather than strict micro-agents and thereby require a subsequent resampling stage. For low-dimensional problems, these methods generally perform well and do not suffer from sparsity. However, because of the rigorous dependency of the original sample, these methods cannot approximate high-dimensional data structures as this will require very large samples. A common problem when using these methods for higher dimensions is therefore the problem of zero-cells, which in addition to the rendering of sparse samples may also lead to convergence and division by zero problems as described in \cite{choupani_2016_ipfReview}. 

While deterministic approaches imply the existence of a ground truth sample, probabilistic simulation-based approaches consider the sample as one of many possible realizations from a ground truth distribution, a distribution that these methods try to approximate in different ways. A common approach is to use conditional feature distributions such as Gibbs sampling \citep{birkin_synthesis_1988, farooq_gibbs_2013} and to approximate the population distribution by drawing iteratively from conditionals. However, this approach suffers from the same shortcomings as the above mentioned deterministic approaches in that the zero-cell problem and the scalability remains to be a challenge. In the statistical literature, it is well acknowledged that the Gibbs approximation of distributions with many features often lead to what is referred to as overfitting \citep{justel_gibbs_1996}. This is essentially caused by invalid marginal distributions when the number of dimensions increases beyond what is supported by the data. This implies that the algorithm will have difficulties escaping the starting solution, and from the perspective of population synthesis we are once again back to the problem of zero-cells. 

In contrast, the most recent model-based simulation approaches, which will be reviewed in more detail in Section~\ref{sec:lit_rev}, can overcome the challenge of scalability but additionally provide a means to estimate probabilities for rare combinations and thereby circumvent the zero-cell problem. This is accomplished by imposing structure for the modelled distribution and for the variables on which it operates. Approaches of this type are used in the statistical disclosure control literature to generate agents that abide to data protection laws (\cite{hu_dirichlet_2018}), but is also applicable to population synthesis. While model-based simulation provides solutions to the zero-cell problem, the added flexibility comes at a cost. The problem is that feature combinations which should have a selection probability of zero may be assigned a positive probability. Such feature combinations are commonly referred to as `structural zeros' (\cite{hu_dirichlet_2018, manrique_discreteMultivariateLatent_2014}). The existence of this problem often leaves the analyst with the task of ruling out combinations which are logically infeasible. Such combinations may include the assigning of driving licenses to small children and that unemployed people travel to work. Both cited papers offer solutions to the structural zero problem, however, they require the analyst to encode the impossible combinations which may not be feasible for applications in high-dimensional data. The problem is one of the challenges that, along with the accelerated use of complex probabilistic models, deserves more attention in the research literature. 

\subsection{Contribution of the paper}
This paper extends the current literature on population synthesis and the spatial micro-simulation literature in the following ways:
\begin{itemize}
    \item It is carefully described and measured how the different models address the zero-cell problem and we propose a set of new metrics to evaluate models in this particular aspect of population synthesis.
    \item Implicit generative models are proposed as a way to synthesize population. Specifically, for the first time, the Wasserstein Generative Adversarial Network (WGAN) \citep{arjovsky_wgan_2017}, is used for this task. 
\item The proposed method applies to high dimensional data and supplements and extents recent work by \cite{borysov_vae_2019} by comparing the performance of the Variational Autoencoder \cite{kingma_vae_2014} with the WGAN and in addition synthesizing unstructured zone data and labour market sector information. 
\end{itemize}

The paper is organized as follows. In Section~\ref{sec:lit_rev}, the literature review is provided. Section~\ref{sec:method} formalizes the problem and introduces the proposed methodology. In Section~\ref{sec:application}, a case study, evaluation procedure, results and discussion are offered. Section~\ref{sec:conclusion} provides conclusion and future work.

\section{Literature review}
\label{sec:lit_rev}

A complete review of all the different population synthesis approaches that have been used historically cannot be justified in the current paper. Rather we will refer to Section 1.2 for a brief description of the different main typologies and in the following section solely focus on the most recent model developments in population synthesis which, by and large, consider the problem from a joint probabilistic perspective.   

A first attempt of using a simulation-based approach for the specific task of population synthesis should be attributed to \cite{farooq_gibbs_2013}. Although the paper is essentially describing a Gibbs sampler implementation, it distinguishes itself from earlier work \citep{birkin_1995_using} by suggesting the use of modelled marginal distributions and partial joint distributions rather than full conditionals derived strictly from the data. Later, in \cite{sun_bn_2015}, a Bayesian network approach for population synthesis was proposed. The idea is to use a Bayesian network model to estimate a probabilistic graph of the data-generation process. Although the model performs well when compared to many other models (IPF and Markov Chain Monte Carlo methods) and tested on a small sample, it remains unclear if the model is applicable to large-scale problems. \cite{sun_mixture_2018} proposed a hierarchical mixture modelling framework for population synthesis that extends their previous work using Bayesian networks. The proposed model assumes multiple latent classes at the household level and multiple latent classes at the individual level for each household latent class. The idea is to model the probability of belonging to a latent class and at the same time model the feature distribution conditional on being in this class. The characteristics follow a multinomial distribution, conditional on a latent class and weighted by a mixture proportion distribution of belonging to that specific class. The model is able to capture marginal distributions of all the variables and the combined joint distributions of all variables. The challenge of this approach is the issue of scalability to many dimensions but also robustness with respect to the selection of the size of the latent classes. Hidden Markov Models (HMM) as presented in \cite{saadi_hmm_2016} represent another model-based approach to population synthesis. The essential idea is that the observed features of certain phenomena are associated to hidden (latent and discrete) states of that phenomenon. As the name suggests, these states are formed in a Markovian process, such that each hidden state is fully determined by its previous state. This is measured by probability tables that determine the transition rates between states. In the HMM paper the idea is to let each state correspond to an attribute. To represent all attributes for a given individual, all attributes are sampled in sequences from the HMM. This process carries on for all attributes and for all individuals. In practice, as the reader might understand, this closely resembles a Gibbs sampling scheme although some additional flexibility can be supported depending on how transition probabilities are modelled. 

Deep generative models have had a recent upsurge in the machine learning literature \cite{goodfellow_dl_2016}. These models are extensions of the probabilistic graphical models presented in the last paragraph but with the use of neural networks as a means to provide more flexible models. Some known examples in the machine learning literature are the Boltzmann machine \cite{ackley_boltzmann_1985}, the Variational Auto-Encoder (VAE) \cite{kingma_vae_2014}, the Generative Adversarial Networks (GAN) \cite{goodfellow_gan_2014}, the normalizing flows \cite{rezende_nf_2015}, and deep autoregressive models, for example, based on Recurrent Neural Networks. In fact, these are the building blocks on which the current generative modelling literature is based. These models have the potential to learn high dimensional distributions by using neural networks as explained in Section~\ref{ssec:NN}. In the context of population synthesis, they have been proposed to synthesize transport data \citep{borysov_vae_2019, borysov_psp_2019} and medical data \citep{choi_medgan_2017, yoon_pategan_2018}. \cite{borysov_vae_2019} propose to use a Variational Auto-Encoder (VAE) to synthesize transport data. The authors find that the model is suitable for generating populations in high dimensional settings and compare it to the Gibbs sampler. \cite{borysov_psp_2019} synthesize panel data from repeated longitudinal studies using a Conditional VAE (CVAE). They find that the CVAE smoothens some of the possible 'holes' in the sparsity of the repeated simulated data. \cite{choi_medgan_2017} generate a population of medical records based on a combined model referred to as the medGAN. The idea of medGAN is to combine a GAN with an autoencoder architecture. \cite{yoon_pategan_2018} propose a GAN with Private Aggregation of Teacher Ensembles (PATE) in order to maintain the privacy of the patients. \cite{godwing_pate_travel_2019} use the same model to share travel diary data with precise latitude and longitude locations. In the last three papers, however, the authors do not explore the zero-cell problem. 

\section{Methodology}
\label{sec:method}


\subsection{Problem formulation}
\label{ssec:problem_formulation}

We consider a population of agents $n=1,...,N$ with each agent defined by a vector of features (variables with the agent's characteristics) $X_n$. Formally, $X$ is a random vector (or a vector of random variables), and each $X_n$ is a realization of this variable. The population synthesis problem is concerned with the estimation of a joint probability distribution $\hat{P}(X)$ that approximates the true joint probability of features across a population, $P(X)$. Provided that this distribution can be reasonably approximated, a population of $N$ individuals can be formed by random sampling from $\hat{P}(X)$. In order to estimate $\hat{P}(X)$, information is typically provided in the form of a sample of size $M < N$ (a survey) that is then used to estimate $\hat{P}(X)$.

The sampling and structural zero problem can be illustrated using three sets. First, a set $\mathcal{X}$ of all possible combinations of values in $X$ that could define an agent. We will refer to this set as the universal set. Second, a set $\mathcal{X}_p$ of all the possible combinations that effectively represent the population. This is the combination of features that appear in the population. Last, the set $\mathcal{X}_s$ which is the feature combinations that exist in the available survey.

\begin{figure}[H]
    \centering
    \includegraphics[width=.5\textwidth]{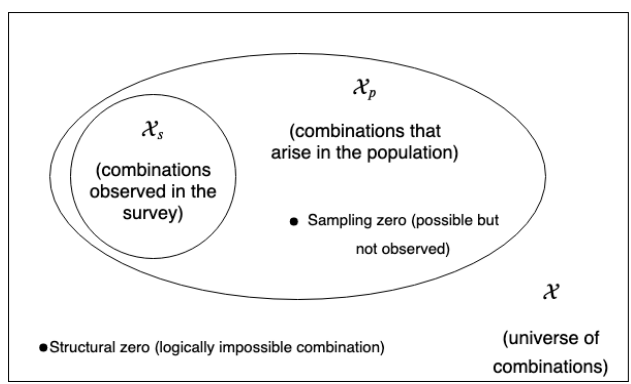}
    \caption{Representation of the sampling zeros, the structural zeros and the different sets of combinations that arise in population synthesis.}
    \label{fig:euler_zeros}
\end{figure}

Naturally, the population $\mathcal{X}_p$ will always include any possible $\mathcal{X}_s$ formed by a survey and we should always have that $\mathcal{X}_s \subseteq \mathcal{X}_p$ across all possible realizations of the survey. A prediction challenge arises if the combination set $\mathcal{X}_p$ is larger than $\mathcal{X}_s$. This is often the case when the number of variables and their categories increases, as the amount of data required to uniformly fill the corresponding hyper-data-cube grows exponentially with the number of dimensions.\footnote{While this is straightforward for $\mathcal{X}$, scaling properties of $\mathcal{X}_p$ might be different (sub-exponential) if the population is not very diverse.} In this case, there are feature combinations that have a non-zero probability of being selected but are not observed. We will refer to these combinations as  ``sampling zeros''. As mentioned in the introduction, these have a close relation to the well-known zero-cell problem in population synthesis. A different type of zeros appears when there are combinations of variables that represent truly impossible combinations, the combinations that are in the universal set of combinations $\mathcal{X}$ but not in the population set of combinations $\mathcal{X}_p$. In the statistical disclosure control literature (\cite{hu_dirichlet_2018}) these are commonly referred to as ``structural zeros''. It is important to stress that impossible combinations may sometimes be better described as very unlikely combinations, for which the modeller chooses to impose some sort of structure to reduce the solution space. One example may be that kids under the age of 15 years are assumed to have zero income. This is not a strictly impossible combination but a way to structure the population in a reasonable way. In Figure~\ref{fig:euler_zeros} a representation of sampling and structural zeros is illustrated using an Euler diagram.

\subsection{Model based approaches}
Complex probabilistic approaches as discussed in the review section can be used for population synthesis. These models require the analyst to impose (or search over) some structure of $P(X)$. The structure comes from establishing conditional dependencies, such as in \cite{sun_bn_2015} and possibly also by introducing latent variables, such as in \cite{sun_mixture_2018, hu_dirichlet_2018, borysov_vae_2019}. These assumptions, if modelled properly, can provide a better estimation of $\hat{P}(X)$.

In this paper we propose using deep generative models for the task of estimating $\hat{P}(X)$. The main idea of these models is to combine probabilistic reasoning and neural networks and it is expected that this type of model can assign probabilities to sampling zeros while also keeping the probability of structural zeros low. In the following we start by describing broadly what a neural network is in the context of population synthesis. Subsequently, we give a formal  presentation of the Generative Adversarial Network model and discuss its properties in this specific context. 

\subsubsection{Neural networks and Variational Auto-Encoders}
\label{ssec:NN}
We can think of a neural network as a very flexible function that links a given input variable to a target variable. It is common to represent this function as a composition of $L$ functions $f_{l}$ where $l=1,\cdots, L$ of an input $x$ mapped to a target variable $y$. That is $\hat{y}=f_{L}(...f_{2}(f_{1}(x)))$. Each of these functions are commonly referred to as ``layers'' in the neural network literature. Particularly, the input variable $x$ is referred to as the input layer,  $f_{l}: \, l=1,...,L-1$ are called ``hidden'' layers and $f_{L}$ the output layer. A neural network is composed of three building blocks: units (which jointly define a layer), edges (weights) and activation functions. The vector $f_{l}$ of single units defines a layer $l$. In the input layer, a numerical variable is represented by one unit while a categorical variable is represented with as many units as number of categories in the variable. The number of hidden layers and units in each hidden layer is defined by the modeller. The matrix $w_{l}$ of edges from layer $l-1$ to layer $l$ represents weights  of a linear combination with which each unit of layer $l-1$ will connect with each unit of layer $l$. Usually, an intercept (also called bias) $b_{l}$ is added to the matrix-vector multiplication $w_{l}f_{l-1}$ Furthermore, the linear combinations defined by the matrix-vector multiplication of the units and the weights are transformed by a point-wise non-linear function $\sigma_{l}$. Putting all these elements together, any layer $f_{l}$ of a neural network is defined by $f_{l}=\sigma_{l}(w_{l}f_{l-1}+b_{l})$.

The training of a neural network is concerned with the estimation of weights that minimize a given loss function. For continuous variables, an example of a loss function would be the mean squared loss while for categorical variables, the loss could be formulated as a cross-entropy function. The most widely used optimization procedures for estimating the weights are gradient-based methods such as stochastic gradient descent and variants of this method. These methods take subsets or ``batches'' of data, estimate the gradient of each parameter with respect to the loss function and average these estimates across the set of observations in order to estimate a single gradient for every parameter to be updated. The gradients of the neural network are estimated using the back-propagation algorithm \citep{rumelhart_bp_1986} which is essentially an application of the chain rule of derivatives. Neural networks have an interesting property that makes them useful for statistical learning and high dimensional population synthesis: they are universal approximating functions \cite{cybenko_approximator_1989, hornik_approximator_1991}. In other words, they can represent arbitrary functions from our input $x$ to our desired output $y$. Beware, however, that this does not imply that the neural network will \textit{generalize} to unseen data. 

\begin{figure}[H]
    \centering
    \includegraphics[width=.99\textwidth]{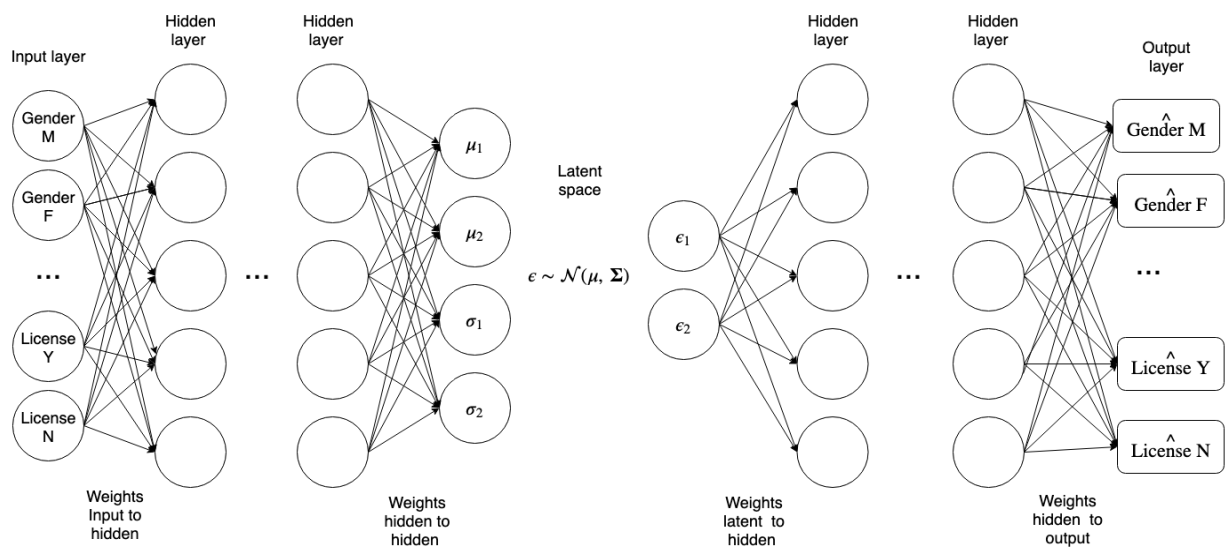}
    \caption{Representation of a VAE.}
    \label{fig:nn_vae}
\end{figure}

One way to tackle the problem of estimating our target population distribution $\hat{P}(X)$ and sampling from it using neural networks is by using the Variational Auto-Encoder (VAE) proposed by \cite{kingma_vae_2014} and \cite{rezende_dgm_2014}. For the population synthesis case, it was proposed by \cite{borysov_vae_2019}. The method is particularly interesting from the perspective of population synthesis as it enables a rendering of agents that are on the one hand diverse, but yet, from a statistical point of view, similar to the agents in the sample data. These features are attained by compressing the feature distribution to a latent space from which the sampling of agents can be based. The latent representation causes a smoothing of the feature space where sampling zeros can be avoided. From a more technical perspective, the VAE consists of two neural networks, a recognition network (or encoder) that maps an observation $x$, in our case all the data that represent an agent, into the parameters of a pre-defined family of distributions, a local latent variable $z$. Then, a second neural network, the generator network (or decoder), that maps a random draw from that distribution (the pre-defined distribution with the estimated parameters) back into the space of $x$. The VAE is trained by minimizing the difference between the input individual and the generated individual and a regularization term which penalizes low entropy which encourages the learned latent distribution to be disperse. When both neural networks are trained, new agents can be sampled by drawing from the prior distribution and subsequently performing a forward pass through the generator. Figure~\ref{fig:nn_vae} depicts a neural network with the Variational Auto-Encoder architecture with socio-economic transport variables. For a deeper understanding of the VAE, we refer the reader to \cite{kingma_vae_2014, doersch_tutorial_2016, kingma_tutorial_2019} and in the context of population synthesis to \cite{borysov_vae_2019}.

\subsubsection{Generative Adversarial Networks}
\label{sssec:GANs}

The family of generative models, known in statistics and machine learning as ``implicit'' generative models \cite{diggle_implicit_1984, mohamed_2016_implicit}, represents a different way of tackling the problem of population synthesis in many dimensions. While the VAE approximates the distribution based on the likelihood of our data, implicit generative models define a stochastic procedure that directly generates data \citep{mohamed_2016_implicit}. This is relevant in the population synthesis problem since we do not need the explicit probability of a certain individual in the population; instead we need to create or simulate agents that resemble those of the population. The way implicit models generate observations is by transforming randomly generated numbers from a latent variable $z \sim p(z)$ (we can think of an $L$-dimensional standard normal) with a deterministic function $G$. We want our function $G \colon \mathbb{R}^{L} \rightarrow X$ to generate observations consistent with $P(X)$ from an $L$-dimensional random variable. 

The fact that neural networks can approximate any function makes them a natural choice for the population synthesis problem, that is, the approximation of the function $P(X)$. A new framework to train these models was introduced in \cite{goodfellow_gan_2014} using what is commonly referred to as adversarial training. The idea of adversarial training is to train two neural networks at the same time. The generator, which in this case is the function $G$, is initiated with a draw from a latent variable $z \sim p(z)$. The draw is then transformed in such a way that the output (agent) is as realistic as possible. The second network, the discriminator $D$ (also called critic), receives an observation, which in our case is an agent. This data can either be from real data or from the generator $G$. The objective of the discriminator is to tell whether the information it receives comes from the real data. In other words, the output of the discriminator $D(x)$ afyer the forward pass of the sample $x$ is the probability of $x$ of coming from the real data, $P(\mathrm{Real}|x)$. The objective of the generator, on the other hand, is to generate samples with the aim of ``fooling'' the discriminator. The name ``adversarial'' then refers to the competition between these two networks. In the process, feedback from the discriminator network is used in the generator network to improve its capability of generating realistic agents. If the discriminator guessed that the sample generated by the generator was likely to be real, the generator does not move much away from that parameter configuration. On the other hand, if the discriminator guessed that the generated sample was likely fake, the generator moves away from that configuration of the parameters. 

An iteration of the training process will proceed as follows. First, the generator takes a random sample from an $L$-dimensional multivariate normal distribution. The random noise is then propagated through the generator network to generate a sample (in a similar way as for the VAE). As an example, consider a person aged 5-15 years and with a driving licence. This generated agent is now used as input in the discriminator network and the role of the discriminator is to predict whether the agent comes from the pool of real samples or not. If the discriminator is well trained it will know that a person aged 5-15 years cannot have a driving licence and the agent will be assigned a low probability of coming from the pool of real samples. Subsequently, both the weights of the discriminator and generator networks are updated using the back-propagation algorithm and a stochastic optimization procedure as explained in Section~\ref{ssec:NN}. The loss function to update both the discriminator and the generator is explained in Equations (\ref{eq:discriminator}) and (\ref{eq:generator}). The training of these models is illustrated in Figure~\ref{fig:GANS_during_training} below.

Formally, the GAN can be derived from a game-theoretic framework. More precisely, the two networks $(D,G)$ play the following Minimax game with value function $V(D,G)$:
\begin{equation}\label{eq:minmax}
\min_G \max_D V(D,G)= \mathbb{E}_{x \sim p_{data}(x)}[\ln D(x)] + \mathbb{E}_{z \sim p_{z}(z)}[\ln (1-D(G(z)))]
\end{equation}
The first term on the right hand side of the equation is the expected value of the $\log$-probability that the discriminator assigns to a sample $x$ from real distribution. This term, as we mentioned before, is then maximized by the discriminator. The second term is one minus the $\log$-probability that the discriminator assigns to an agent generated by $G$ of being real. Since, in the eyes of the discriminator, this probability should be low, the discriminator seeks to maximize this term while, on the other hand, the generator will seek to minimize the term.  

Equation~(\ref{eq:minmax}) gives the basis for the loss functions that can be maximized simultaneously using an optimization procedure. For a single point $i$ in our data set, the discriminator we want to minimize the loss $L_{D}$
\begin{equation}\label{eq:discriminator}
L_{D} = -\big[\ln D(x^{(i)}) + \ln(1-D(G(z^{(i)})))\big]
\end{equation}
with respect to the discriminator parameters in the neural network, while for the generator we want to minimize the loss $L_{G}$
\begin{equation}\label{eq:generator}
L_{G} = \ln(1-D(G(z^{(i)})))
\end{equation}
with respect to the generator parameters. In practice, however, more updates of the discriminator are performed for each update of the generator.

In this paper, rather than applying the GAN in its basic form, we apply an alternative form referred to as the Wasserstein GAN (WGAN). This model was recently introduced in \cite{arjovsky_wgan_2017} and is proven to have properties that makes it more desirable for generation tasks. Particularly, \cite{arjovsky_wgan_2017}, the losses of the generator and the discriminator are based on a distance metric called the Wasserstein-1 (or ``earth mover's'') distance. When using this distance metric, \cite{arjovsky_wgan_2017} proved that if the generator $G$ is continuous in $\theta$ and locally Lipschitz, the Wasserstein-1 distance is continuous everywhere and differentiable almost everywhere. Under the same conditions, commonly used divergences such as the Kullback-Leibler divergence or the Jensen-Shannon divergence do not hold the same properties. This makes the Wasserstein-1 distance more desirable in an optimization procedure. The practical difference between these and the normal GAN comes in the loss function and from clipping the weights (limiting their size) to force the generator to be Lipschitz. The functions optimized in the WGAN are as follows. For the discriminator:
\begin{equation}\label{eq:wgan_discriminator}
L_{D} = -\big[D(x^{(i)}) + (1-D(G(z^{(i)})))\big]
\end{equation}
and the generator:
\begin{equation}\label{eq:wgan_generator}
L_{G} = 1-D(G(z^{(i)}))
\end{equation}

We refer the interested reader to the original paper for more details about the theoretical basis of changing the loss function. 


\begin{figure}[H]
    \centering
    \includegraphics[width=.99\textwidth]{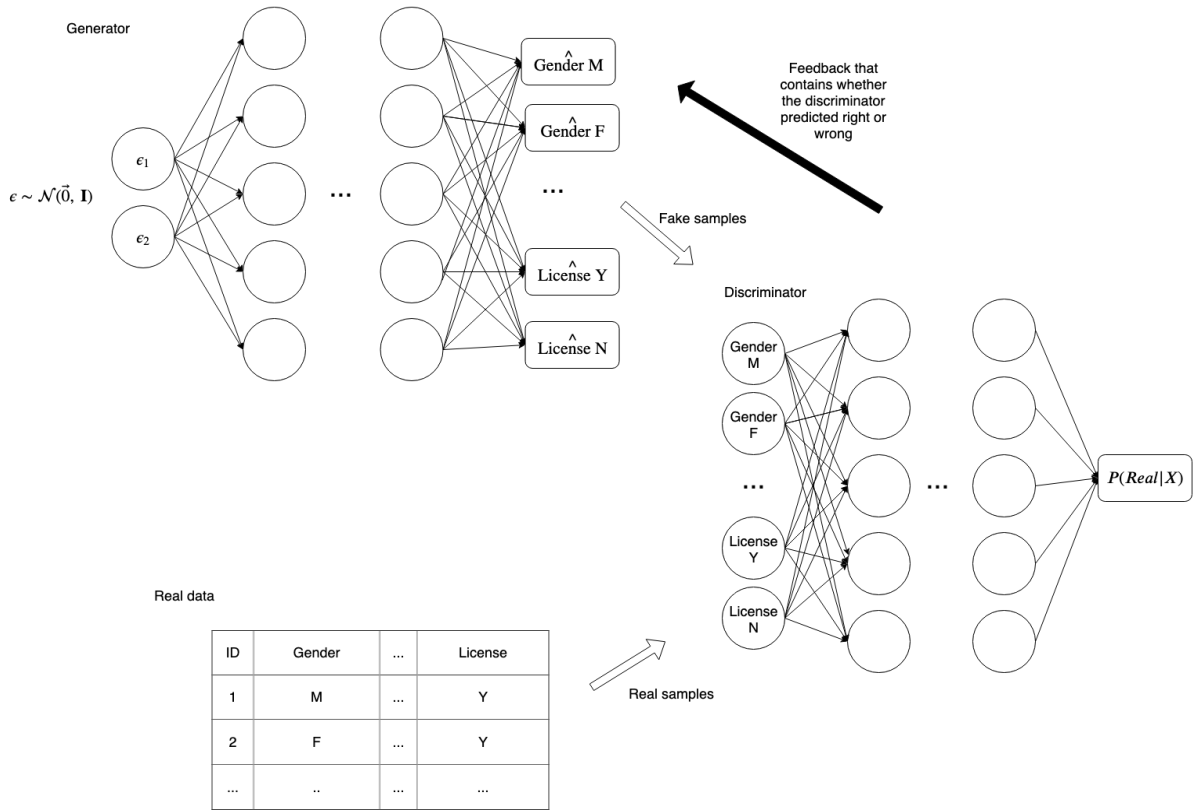}
    \caption{Graphical depiction of GAN. During training, the generator is fed with random numbers from a distribution and generates ``fake'' samples. Then both fake samples and real samples are fed into the discriminator which tries to tell where they come from. This information is fed back into both the generator and the discriminator to improve their parameters.}
    \label{fig:GANS_during_training}
\end{figure}

\section{Application}
\label{sec:application}

\subsection{Data}
\label{ssec:data}

This study is based on data from the Danish National Travel Survey (TU) \cite{tu}. It contains 156,248 individuals from 2006 to 2018. The variables include socio-economic characteristics, such as income, type of dwelling, number of persons in household, etc., and variables related to trips, such as municipality of origin, total time spent on mode, etc. All the variables used in the modelling exercise, their type and their respective number of categories (bins) can be found in Table~\ref{tab:data}. The data is further enriched by job sector information based on data from Statistics Denmark. While not all of the variables might be of interest for a particular type of analysis, the complete data set provides a challenging case when exploring how high-dimensional data can be approximated. 

The data pre-processing included: i) transforming numerical variables into categorical variables which can take one of the 5 values based on the corresponding quantiles, ii) removing the variables with more than 20\% of missing values, iii) the merging of job locations and sectors from Statistics Denmark, iv) dividing the data set into 3 subsets: training (40\%), validation (40\%) and test (20\%) set. After the pre-processing, the data consisted of 67 variables with 30,349 observations for training, 30,349 observations for validation, and 14,054 samples for testing.

\begin{table}[ht!]
    \footnotesize
    \centering
        \begin{tabular}{lllll}
        \hline
        \# & Name & Type & Number of values & Description \\
        \hline
        \hline 
	    1 & DayJourneyType & categorical & 7 & Journey type of the day \\
        2 & DayNumJourneys & numerical (int) & 16 & Number of journeys during 24 hours \\
        3 & DayPrimTargetPurp & categorical & 28 & Primary stay of the day, purpose \\
        4 & DayStartJourneyRole & categorical & 3 & Start of the day: position in journey \\
        5 & DayStartPurp & categorical & 26 & Purpose at start of the day \\
    	6 & DiaryDayType & categorical & 6 & Number of persons in the household \\
    	7 & DiaryMonth & numerical (int) & 12 & Number of persons in the household \\
    	8 & DiaryWeekday & categorical & 7 & Number of persons in the household \\
    	9 & FamNumAdults & categorical & 11 & Number of adults in the family \\
    	10 & FamNumDrivLic & numerical (int) & 12 & Number of persons with a driving licence in the family \\
        11 & Handicap & categorical & 3 & Handicap \\
    	12 & HomeAdrCitySize & numerical (int) & 14 & Home, town size \\
    	13 & HomeAdrNUTS & categorical & 11 & Region and sub-region of the household in Denmark \\
    	14 & HomeParkPoss & categorical & 20 & Parking conditions at home \\	
            15 & HousehAccomodation & categorical & 7 & Home, type \\
            16 & HousehAccOwnorRent & categorical & 4 & Home, ownership \\
    	17 & HousehCarOwnership & numerical (int) & 14 & Car ownership in the household \\
            18 & HousehNumcars & numerical (int) & 15 & Number of cars in the household \\
    	19 & HousehNumPers1084 & numerical (int) & 5 & Number of persons 10-84 years in the household \\
            20 & HwDayspW & numerical (int) & 4 & Number of commuter days \\  
            21 & IncRespondent2000 & numerical (int) & 12 & The respondent’s gross income \\
    	22 & JstartType & categorical & 5 & Journey base, type \\
            23 & ModeChainTypeDay & categorical & 14 & Transport mode chain for the entire day \\	
    	24 & MunicipalityDest & categorical & 99 & Number of persons in the household \\
    	25 & MunicipalityOrigin & categorical & 99 & Number of persons in the household \\
    	26 & NightsAway & numerical (int) & 4 & Number of nights out \\
    	27 & NuclFamNumAdults & numerical (int) & 8 & Number of adults in nuclear family \\
    	28 & NuclFamNumDrivLic & numerical (int) & 7 & Number of persons with a driving licence in nuclear family \\
        29 & NuclFamType & categorical & 5 & Nuclear family type \\   
        30 & NumTripsCorr & numerical (int) & 7 & Number of trips \\	
    	31 & NumTripsExclComTrans & numerical (int) & 7 & Number of trips, without commercial transport \\
    	32 & PopSocio & categorical & 2 & Respondent's occupation \\	
    	33 & PosInFamily & categorical & 5 & Position in the nuclear family \\ 
            34 & PrimModeDay & categorical & 23 & Primary mode of transport for the entire day \\
    	35 & PseudoYear & numerical (int) & 9 & Year of the interview \\  
    	36 & RespAgeCorrect & numerical (int) & 14 & Age \\
            37 & RespEdulevel & categorical & 7 & Educational attainment \\   
    	38 & RespHasBicycle & categorical & 3 & Bicycle ownership \\
            39 & ResphasDrivlic & categorical & 5 & Driving licence \\    
    	40 & RespHasSeasonticket & categorical & 2 & Season ticket (public transport) \\
            41 & RespIsmemCarshare & categorical & 3 & Member of car sharing scheme \\  
    	42 & RespSex & binary & 2 & Gender \\ 
    	43 & Sector & categorical & 36 & Labour market sector of the respondent \\
    	44 & TotalLenExclComTrans & numerical (cont) & 13 & Total travel distance without commercial transport \\
    	45 & TotalMin & numerical (cont) & 13 & Total duration of trips \\
        46 & TotalMotorLen & numerical (cont) & 10 & Total motorized travel distance \\
        47 & TotalMotorMin & numerical (cont) & 10 & Total motorized duration of trips \\
        48 & TotalNumTrips & numerical (int) & 7 & Total number of trips \\
    	49 & WorkHoursPw & numerical (int) & 6 & Number of weekly working hours \\
        50 & WorkHourType & categorical & 5 & Planning of working hours \\
        51 & WorkParkPoss & categorical & 12 & Parking conditions at place of occupation \\
        52 & WorkPubPriv & categorical & 4 & Public- or private-sector employee \\
    	53 & Year & numerical (int) & 8 & Year of the interview \\
        \hline
        \end{tabular}
    \caption{Individual attributes of the TU participants used in the paper. For the numerical variables, the third column denotes the number of categories used when they are converted into categorical variables.}
    \label{tab:data}
\end{table}

Clearly, the number of dimensions for the entire data set makes any use of the IPF algorithm impossible as the number of the corresponding matrix cells grows exponentially with the number of dimensions. For example, considering $n$ binary variables requires fitting of $2^n$ cells that quickly makes the algorithm impractical for large $n$. A data driven simulation approach, such as a Gibbs sampler, will also be infeasible for this type of problem as the empirical marginal distributions will be invalidated. Hence, for this type of problem only model-based simulation approaches apply. In addition, a Bayesian network model would require some assumptions from the researcher, as searching through all the possible graphs spanned by these variables is, again, impractical.

\subsection{Model selection and results}

The validation of generative models, and specifically deep generative models applied to high-dimensional data, is not an easy task. A naive approach to estimating the quality of the generated population is to estimate the empirical joint distribution over all the variables and compare it to the empirical distribution over all the variables of the test set. However, since we are working with a high number of dimensions, this quickly becomes unreliable due to the data sparsity. An alternative way is to validate the quality of the approximated distribution by comparing it with partial versions of the original distributions. For example, one can assess the quality of the joint distribution for three specific variables chosen arbitrarily (say, municipality of origin, municipality of destination and labour market sector) and estimate the corresponding empirical joint distribution of these variables and finally compare them using some metric. Here the partial joint distributions of the variables $i\ldots j$ will be denoted as $\pi_{i\ldots j}$ for the test set and $\hat{\pi_{i\ldots j}}$ for the generated set. Following the literature on population synthesis \citep{muller_2011_populationSynthesisSOTA, pritchard_sparsity_2012, farooq_gibbs_2013}, we use the Standardized Mean Squared Error (SRMSE) metric to compare these empirical distributions. The idea is to calculate the Root Mean Squared Error (RMSE) for both distributions and divide it by the sum of the test probabilities divided by the number of different possible combinations denoted by $N_c$. Since the joints' probabilities are empirical distributions, this sum should add to 1 and thus, the SRMSE is calculated by multiplying the RMSE by the square root of the number of possible combinations of that distribution:

$$
\operatorname{SRMSE}(\hat{\pi}, \pi)=\frac{\operatorname{RMSE}(\hat{\pi}, \pi)}{\overline{\pi}}=\frac{\sqrt{\sum_{i} \cdots \sum_{j}\left(\hat{\pi}_{i \ldots j}-\pi_{i \ldots j}\right)^{2} / N_c}}{\sum_{i} \cdots \sum_{j} \pi_{i \ldots j} / N_c}=\sqrt{\sum_{i} \cdots \sum_{j}\left(\hat{\pi}_{i \ldots j}-\pi_{i \ldots j}\right)^{2}N_c}
$$

\begin{figure}[H]
    \centering
    \begin{subfigure}[b]{\textwidth}
        \includegraphics[width=.33\textwidth]{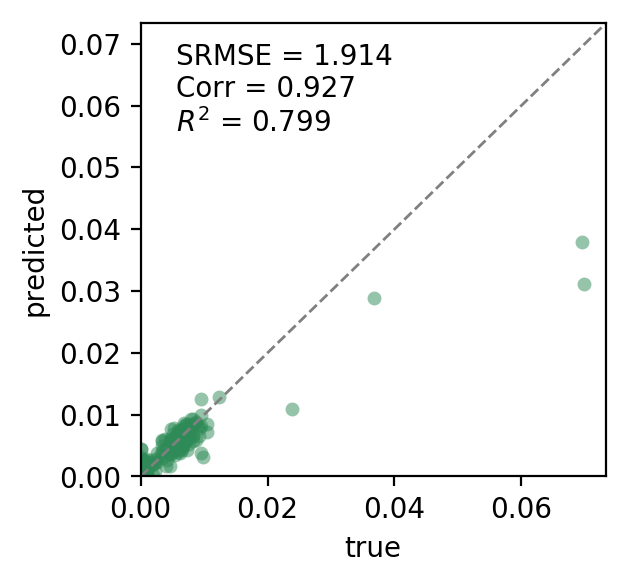}\hfill
        \includegraphics[width=.33\textwidth]{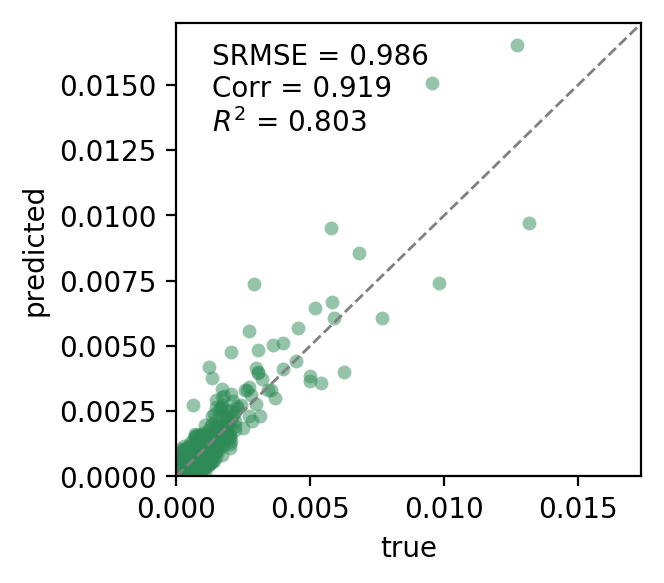}\hfill
        \includegraphics[width=.33\textwidth]{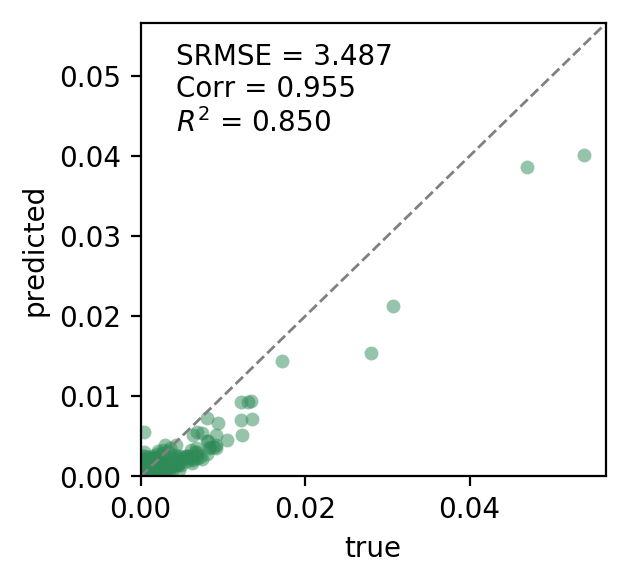}
        \caption{Variational Auto-Enocoder}
        \label{fig:vae_triptic}
    \end{subfigure}
    
    \centering
    \begin{subfigure}[b]{\textwidth}
        \includegraphics[width=.33\textwidth]{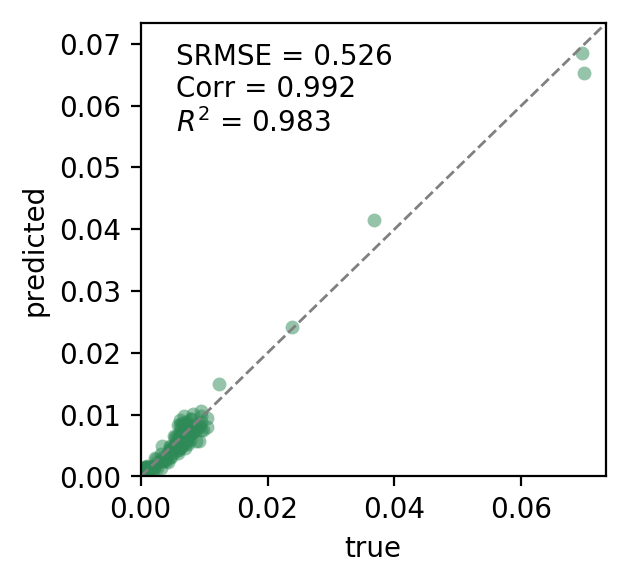}\hfill
        \includegraphics[width=.33\textwidth]{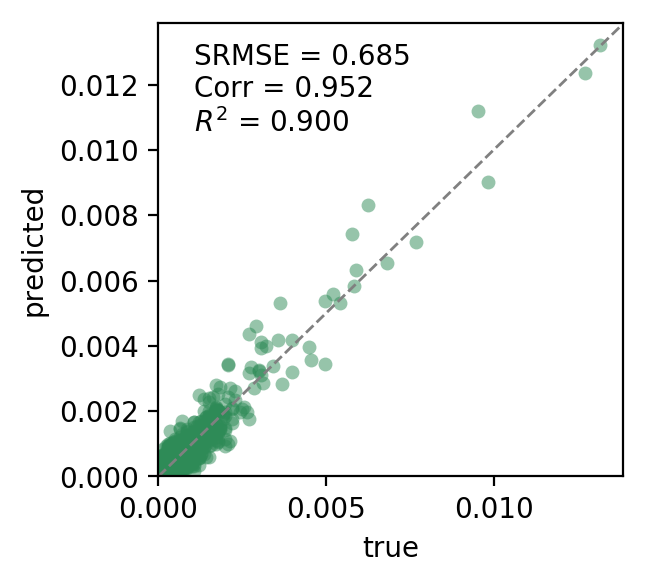}\hfill
        \includegraphics[width=.33\textwidth]{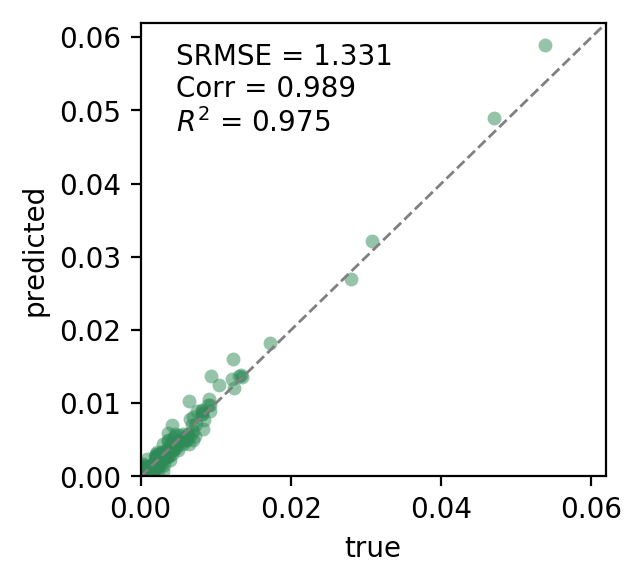}
        \caption{Wasserstein Generative Adversarial Network}
        \label{fig:wgan_triptic}
    \end{subfigure}
    
   \caption{Joint distribution of income, driving license, age and occupation (left), municipality of origin and labor sector (center) and municipality of origin and municipality of destination (right) for the VAE (upper row) and the WGAN (lower row). The figures represent scatter plots of the true and approximated distributions, where bin frequencies are plotted on the horizontal and the vertical axis, respectively. The SRMSE, the Pearson correlation coefficient and the $R^{2}$ are shown in the top left corner of each graph.}
   \label{fig:vaevswgan}
\end{figure}

We also compare both distributions using the Pearson correlation coefficient and the coefficient of determination $R^{2}$ which are defined in the usual way for the same bin frequencies. In order to provide a more qualitative assessment of the different models, several ``45-degree''-plots are provided in Figure~\ref{fig:vaevswgan}. In these plots, different subsets of variables are selected and evaluated in a  similar way as for the partial distributions described above. From the figures, it can be seen that the WGAN performs better when compared to the VAE. In Figure~\ref{fig:very_high_dim} we showcase the performance of both models when stress-tested for higher dimensions. The figure represents the same distribution matching for municipality of origin, which contains 99 different classes, municipality of destination, with 99 classes, 12 income classes and 36 labour market sectors. In total, these four variables represent more than four million possible combinations. It is possible to see that, even though the distributions are not matched perfectly, and in particular for sparse combinations, the overall correlation pattern is generally well represented. As before, the WGAN (center) provides a better approximation as compared to the VAE (left). As an additional test, we also compare the distribution for the training data with the approximated distribution from the test data (Right panel in Figure~\ref{fig:very_high_dim}) as an alternative way of benchmarking the deep generative models. Again, from a validation perspective, the WGAN approximation is better when compared to the VAE in this case. The SRMSE for the training data, however, is lower for the training set while the Pearson correlation coefficient and the $R^2$ are better for the WGAN .

Following the standard machine learning methodology, the training set is used to fit the model parameters, while the free parameters of the model (also called hyper-parameters in the machine learning literature) are chosen based on the performance on the validation set using a gradient-free optimization procedure called Bayesian optimization \cite{frazier_BOtutorial_2018}. The hyper-parameters of the model in our case are the neural network architectures: number of layers, number of units per hidden layer, dimension of latent space, etc., and the optimization parameters, the optimizer, the learning rate, etc. Finally, the best-performing model is evaluated using the test set.

The WGAN is compared against the Variational Autoencoder  (VAE) \cite{kingma_vae_2014} which was proposed as a population synthesizer in \cite{borysov_vae_2019}, a generator that uses the marginals to generate observations and a purely random generator.

\begin{figure}[H]
    \includegraphics[width=.33\textwidth]{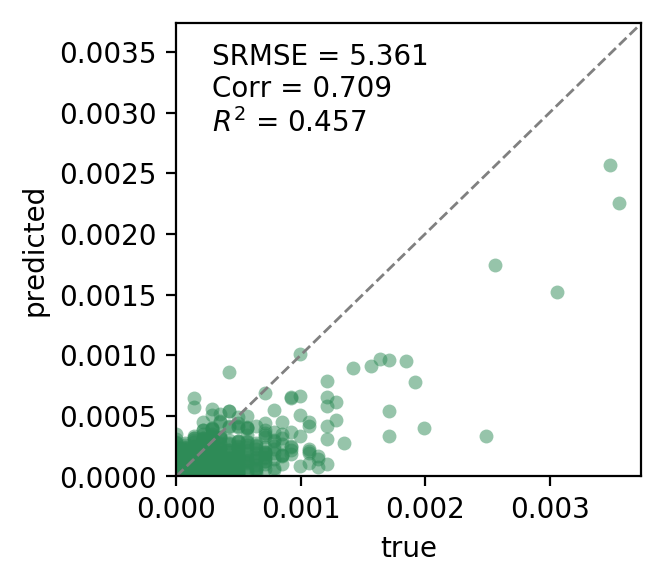}\hfill
    \includegraphics[width=.33\textwidth]{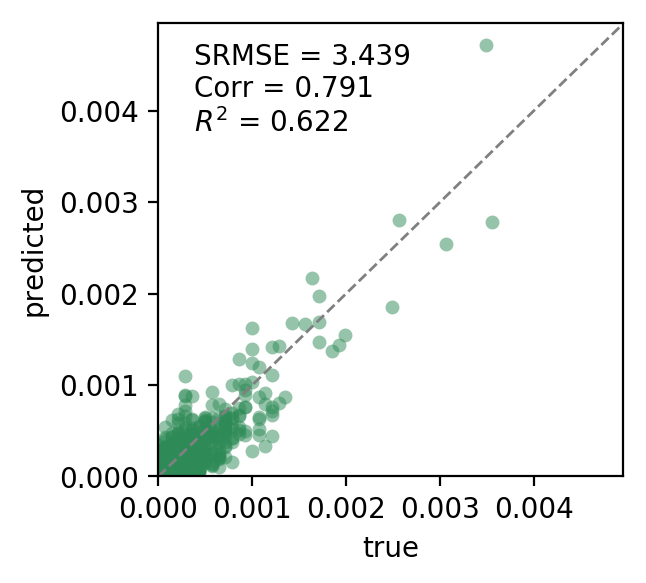}\hfill
    \includegraphics[width=.33\textwidth]{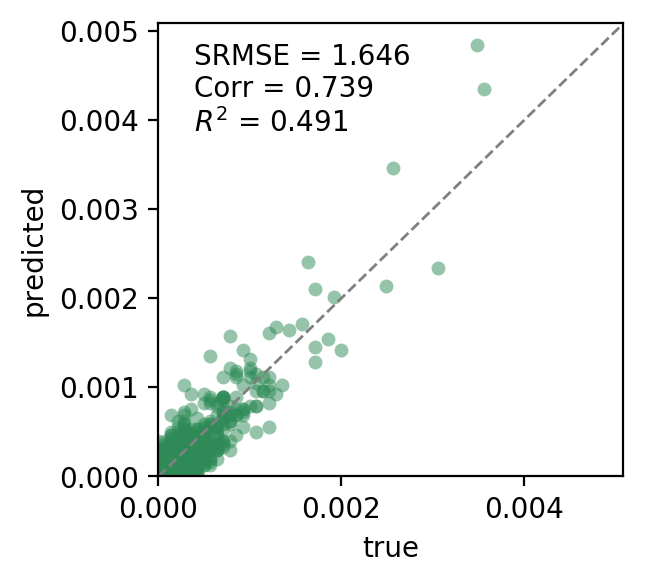}\hfill
    \caption{High dimensional joint distribution matching for municipality of origin (99 levels), municipality of destination (99 levels), income of the respondent (12 levels) and labour market sector (36 levels). VAE on the left side, WGAN in the center and training data on the right side.}
    \label{fig:very_high_dim}
\end{figure}

\subsection{Sampling zeros validation}
\label{ssec:model_eval}

The main objective of the paper is to investigate the ability of these models to recover sampling zeros. Since the true sampling zeros are not available in the data by definition, we artificially create them. For this purpose, we define sampling zeros as combinations of variables which are in the test set but not in the training set. After the distribution has been approximated, we examine how many sampling zeros are recovered using the models as a proportion of the total sampling zeros. This metric should approach 1 for good models. However, since the models are generating individuals who are not in the training data set, there is also some probability that some of these individuals will not appear in the test data. These are the individual we call structural zeros in Section~\ref{ssec:problem_formulation}. Ideally, this metric should approach zero for good models. Both of these measures are useful in the assessment of the zero-cell problem and offer a different perspective with respect to the performance of these models. While, as mentioned before, such indicators only represent a partial validation of the models, we believe that using the ratio between these two metrics serves as a good way of validating the models. This ratio represents the number of sampling zeros generated for each structural zero, and the lower the metric, the better the model. We plot the ratio as a function of the number of observations generated. This allows us to see how the ratio evolves as we generate individuals. As baseline models, we use sampling attributes independently from marginal distributions and as well as generating the attributes completely at random from uniform priors.

In Figure~\ref{fig:sampling_structural_ratio}, this metric is plotted for three cases of different dimensionality: \~10k-dimensional joint (a), \~2.5M-dimensional joint (b), and  \~840M-dimensional joint (c). The first row of Figure~\ref{fig:low_dim} depicts the case which includes all the possible combinations of municipality of origin and municipality of destination. For these variables, the number of sampling zeros generated between the training and test split was 325. The plot shows that a randomly generated sample as well as a sample generated from pure marginal distributions is able to recover all the sampling zeros while the VAE and the WGAN are not able to do so. On the other hand, both the random model and marginal model generate more structural zeros than the deep generative models. The ratio is more favourable for the latter models with the WGAN performing better than the VAE. Figure~\ref{fig:med_dim} and Figure~\ref{fig:high_dim} show the same metrics for the second case of 2.5M and the third case of 840M combinations, respectively. In the second case, the random and the marginal generators are worse at generating sample zeros compared to the VAE and the marginal generator has a performance similar to the WGAN when generating sampling zeros. 
One interesting thing to note from this graph is that if we generate enough samples, we could create all the sampling zeros with the random or marginal generators, however, this will come result in an similar generation of structural zeros. 
For all the cases, the WGAN performs the best according to the ratio introduced.

\begin{figure}[H] \ContinuedFloat
    \centering
    \begin{subfigure}[b]{\textwidth}
        \includegraphics[width=.33\textwidth]{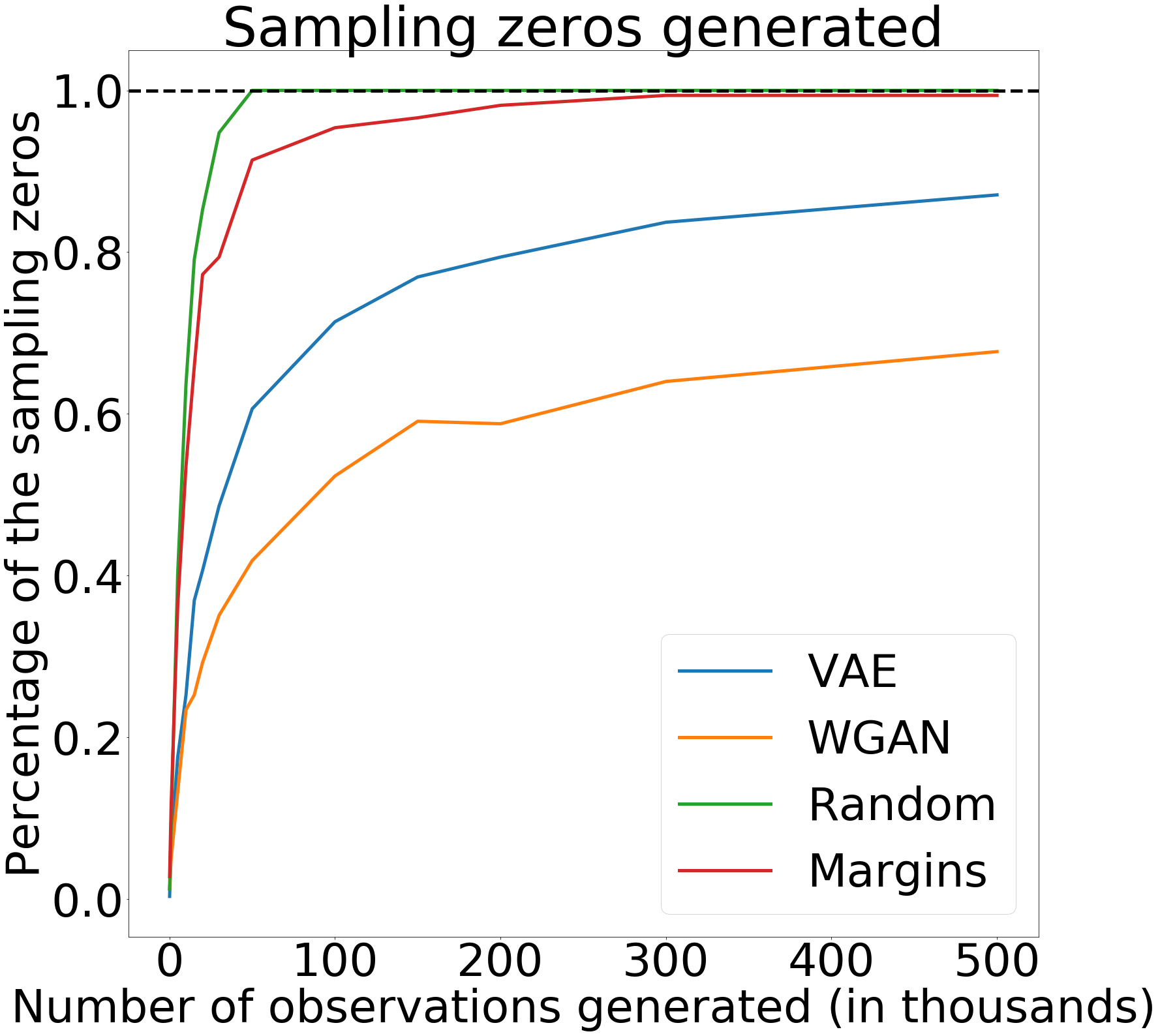}\hfill
        \includegraphics[width=.33\textwidth]{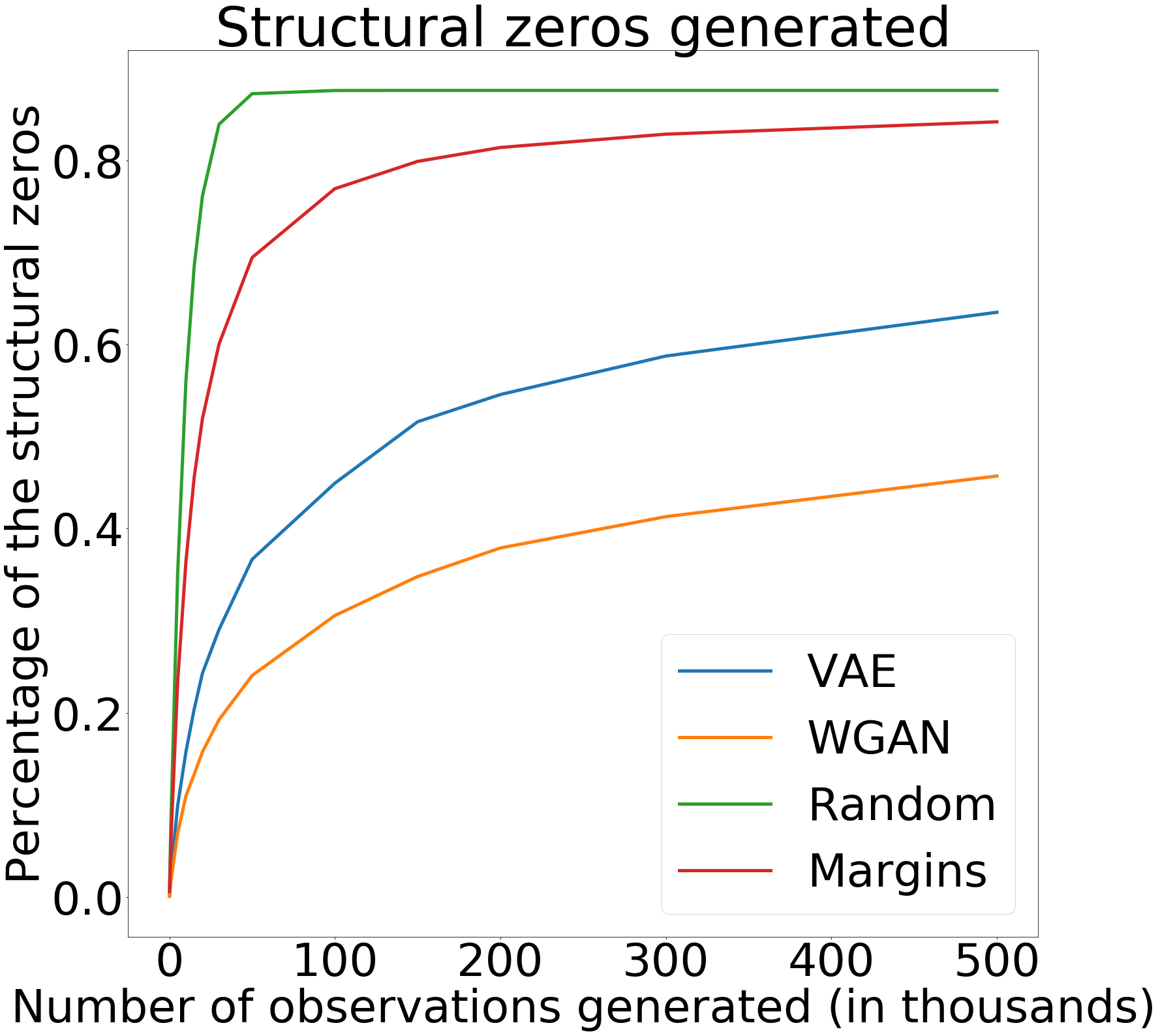}\hfill
        \includegraphics[width=.33\textwidth]{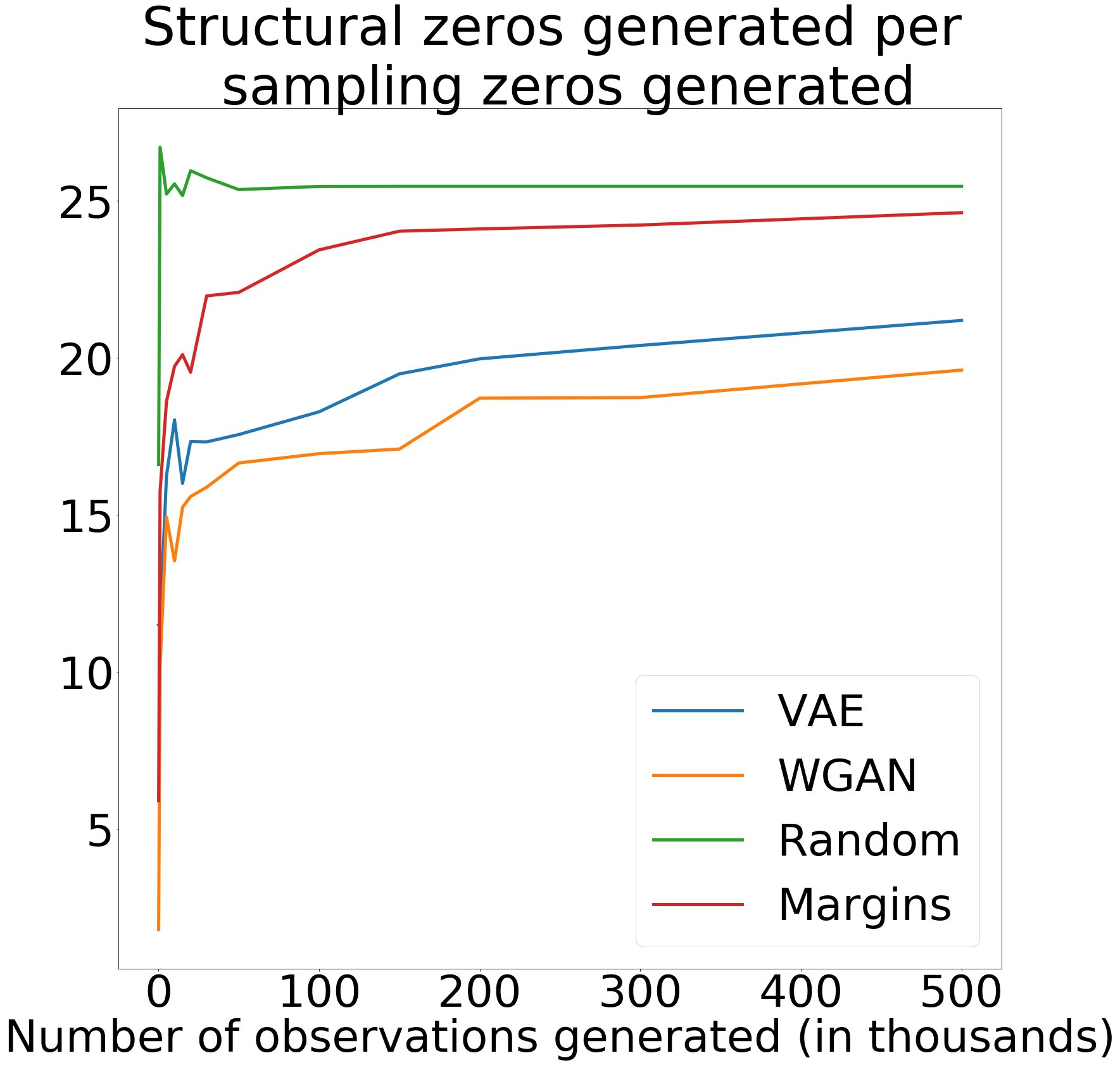}
        \caption{Municipality of origin (99 levels), Municipality of destination (99 levels); 9801-dimensional joint}
        \label{fig:low_dim}
    \end{subfigure}
    
    \centering
    \begin{subfigure}[b]{\textwidth}
        \includegraphics[width=.33\textwidth]{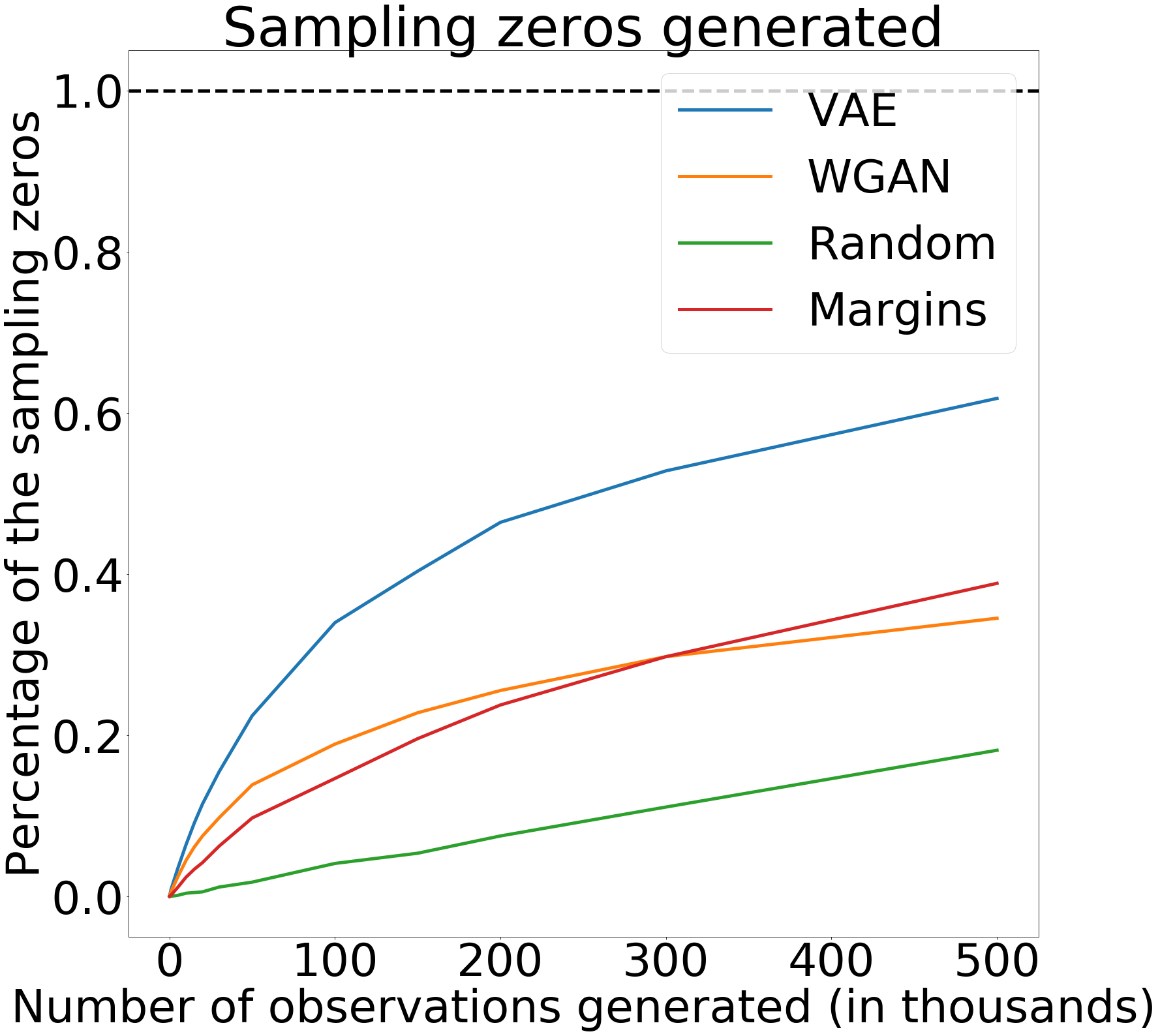}\hfill
        \includegraphics[width=.33\textwidth]{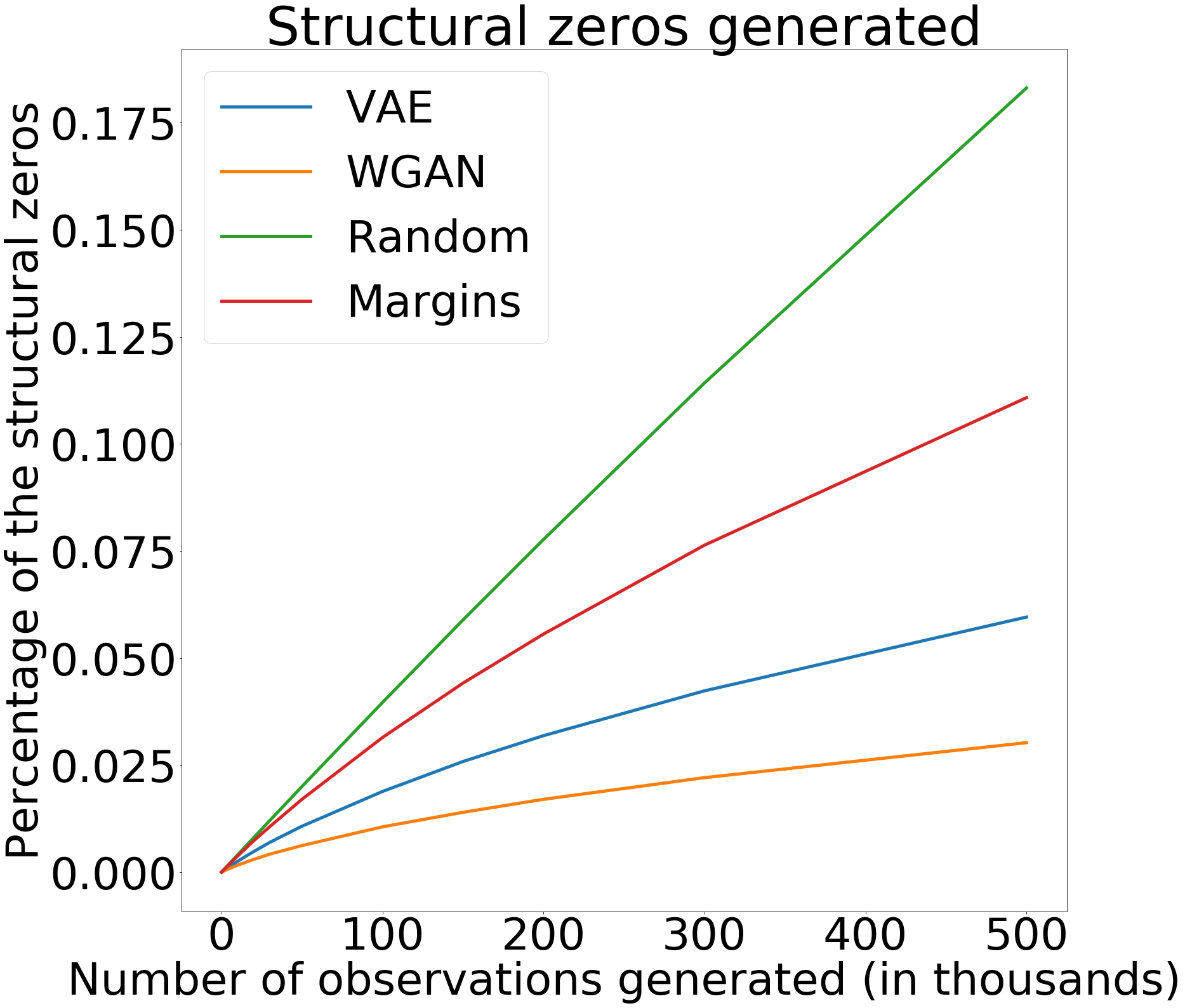}\hfill
        \includegraphics[width=.33\textwidth]{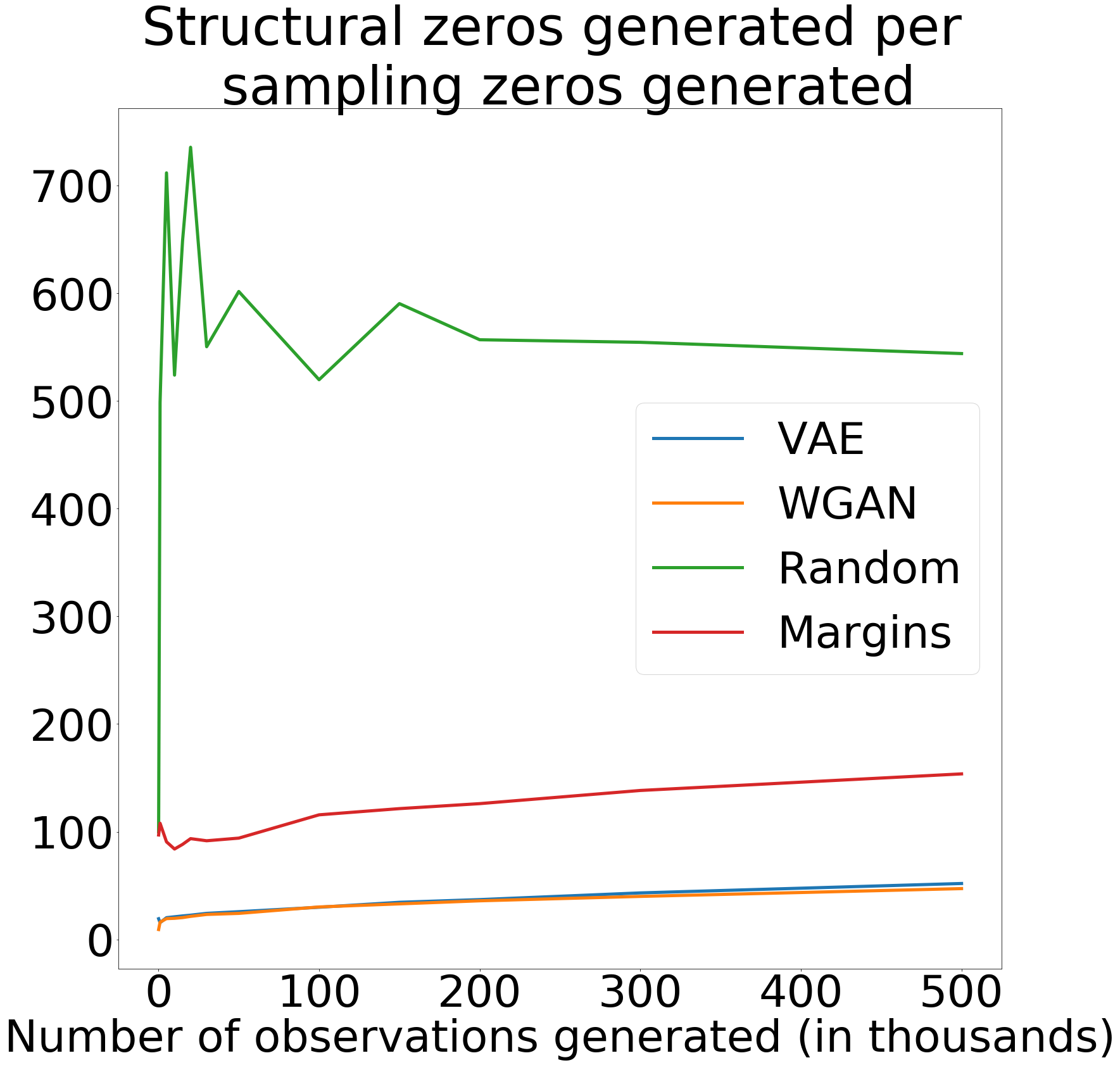}
        \caption{Municipality of Origin (99 levels), Municipality of Destination (99 levels), Sector (36 levels), Education (7 levels); \~2.5M-dimensional joint}
        \label{fig:med_dim}
    \end{subfigure}
    
    \centering
    \begin{subfigure}[b]{\textwidth}
        \includegraphics[width=.33\textwidth]{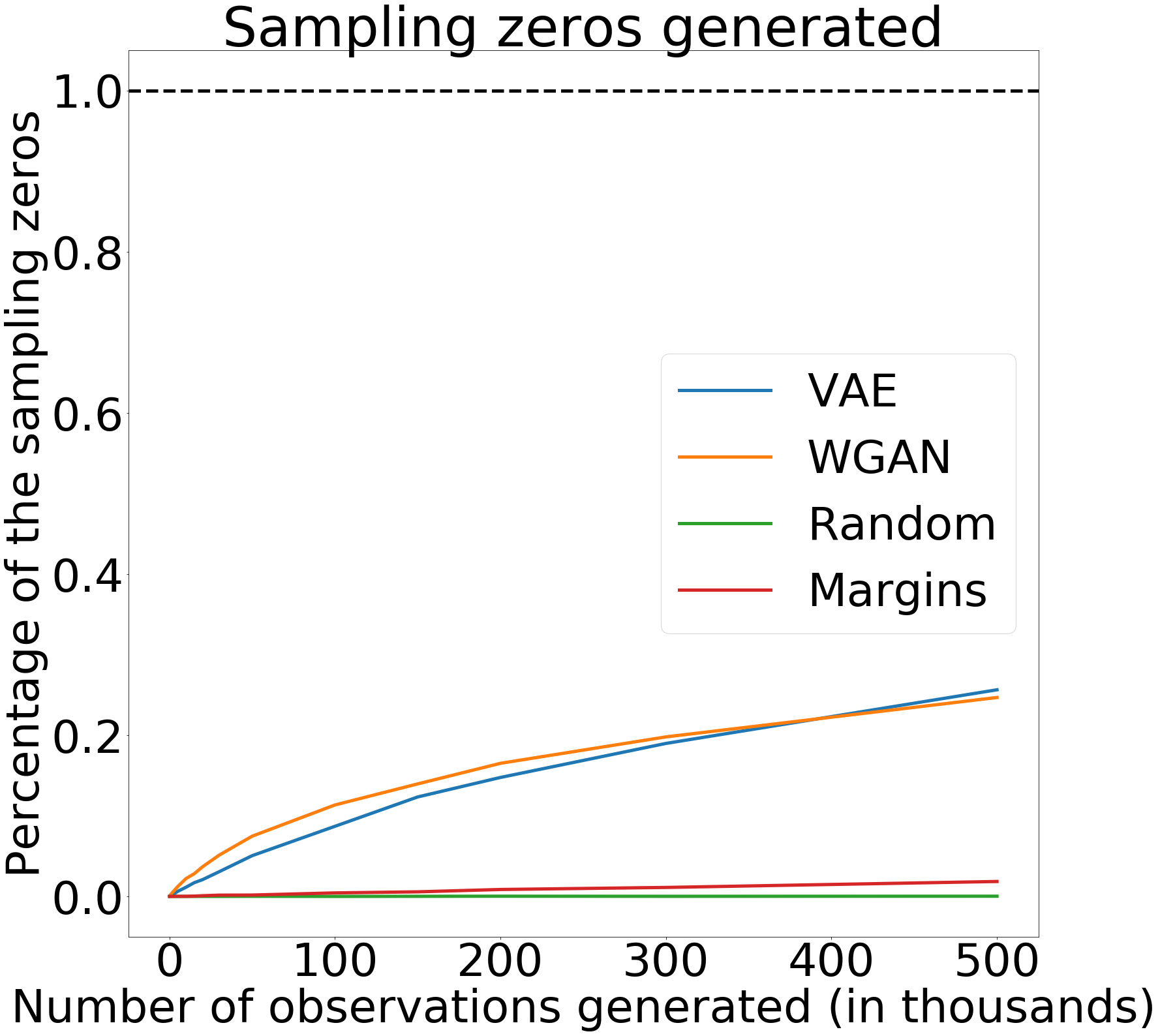}\hfill
        \includegraphics[width=.33\textwidth]{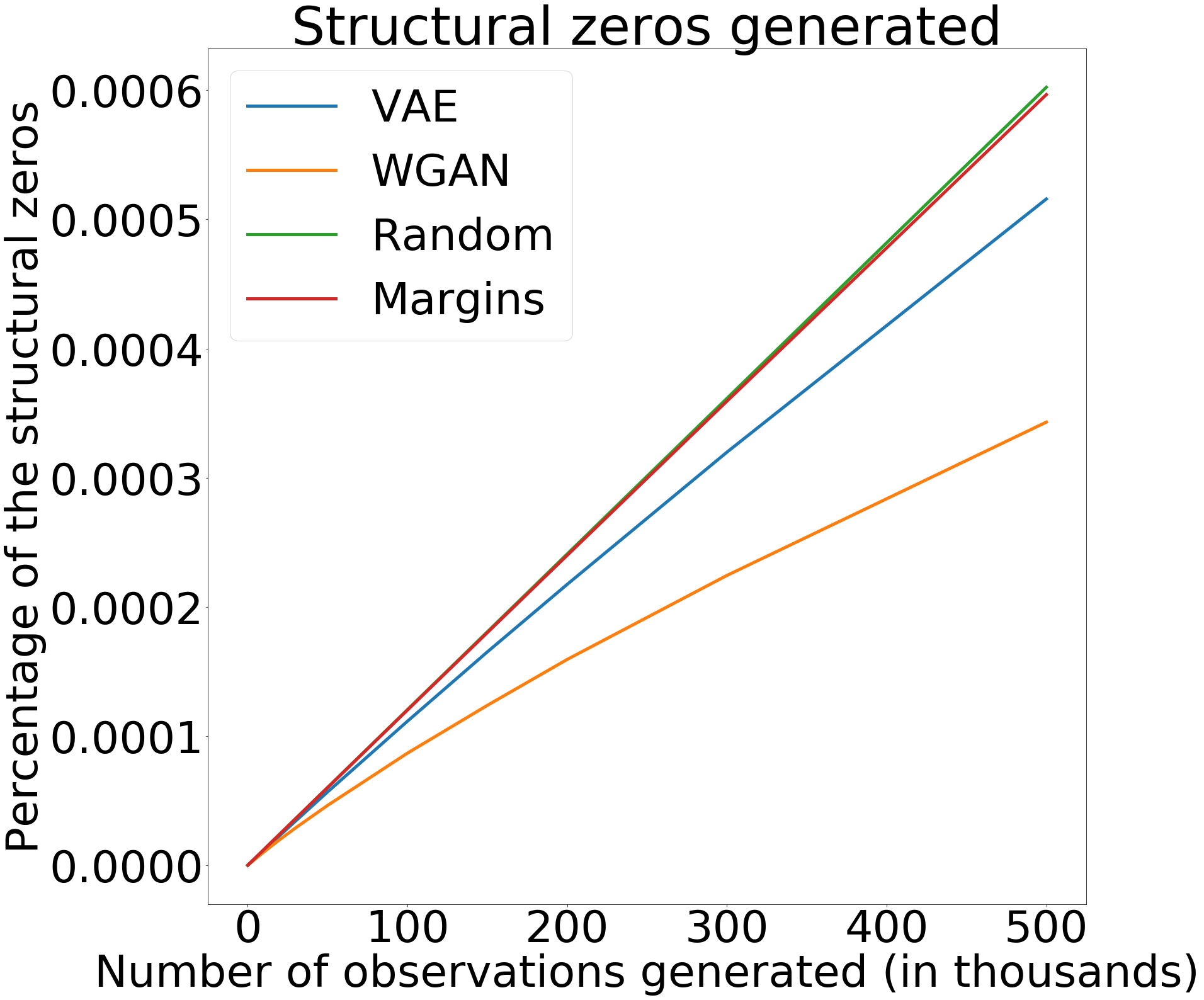}\hfill
        \includegraphics[width=.33\textwidth]{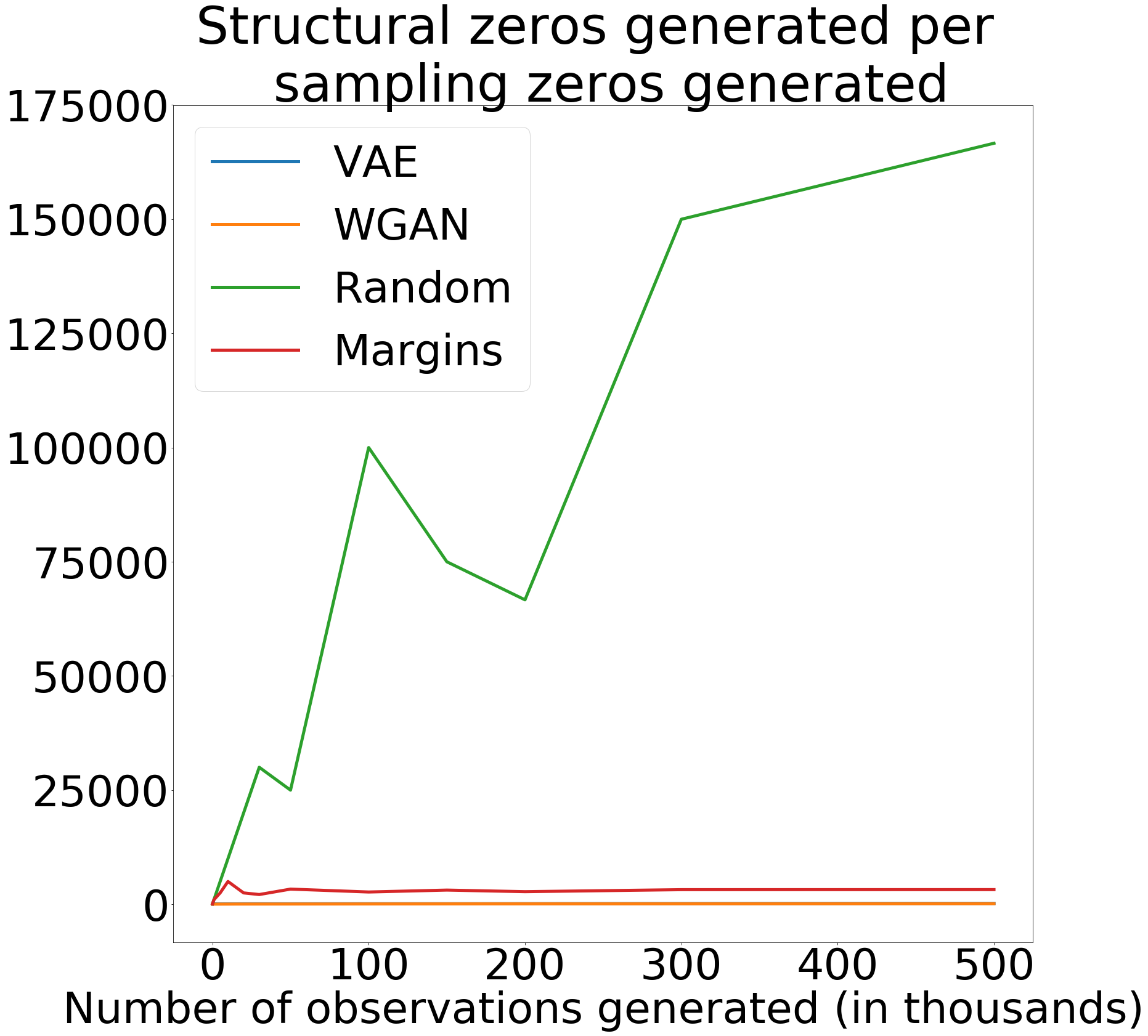}
        \caption{Municipality of Origin (99 levels), Municipality of Destination (99 levels), Sector (36 levels), Education (7 levels), Income (12 levels), Age (14 levels), Gender (2 levels); \~840M-dimensional joint}
        \label{fig:high_dim}
    \end{subfigure}
    
   \caption{Ratio between the number of sampling and structural zeros produced by the different models for different subsets of variables with different dimensionality.}
   \label{fig:sampling_structural_ratio}
\end{figure}

\begin{figure}[H]
    \centering
    \begin{subfigure}[b]{\textwidth}
        \includegraphics[width=.33\textwidth]{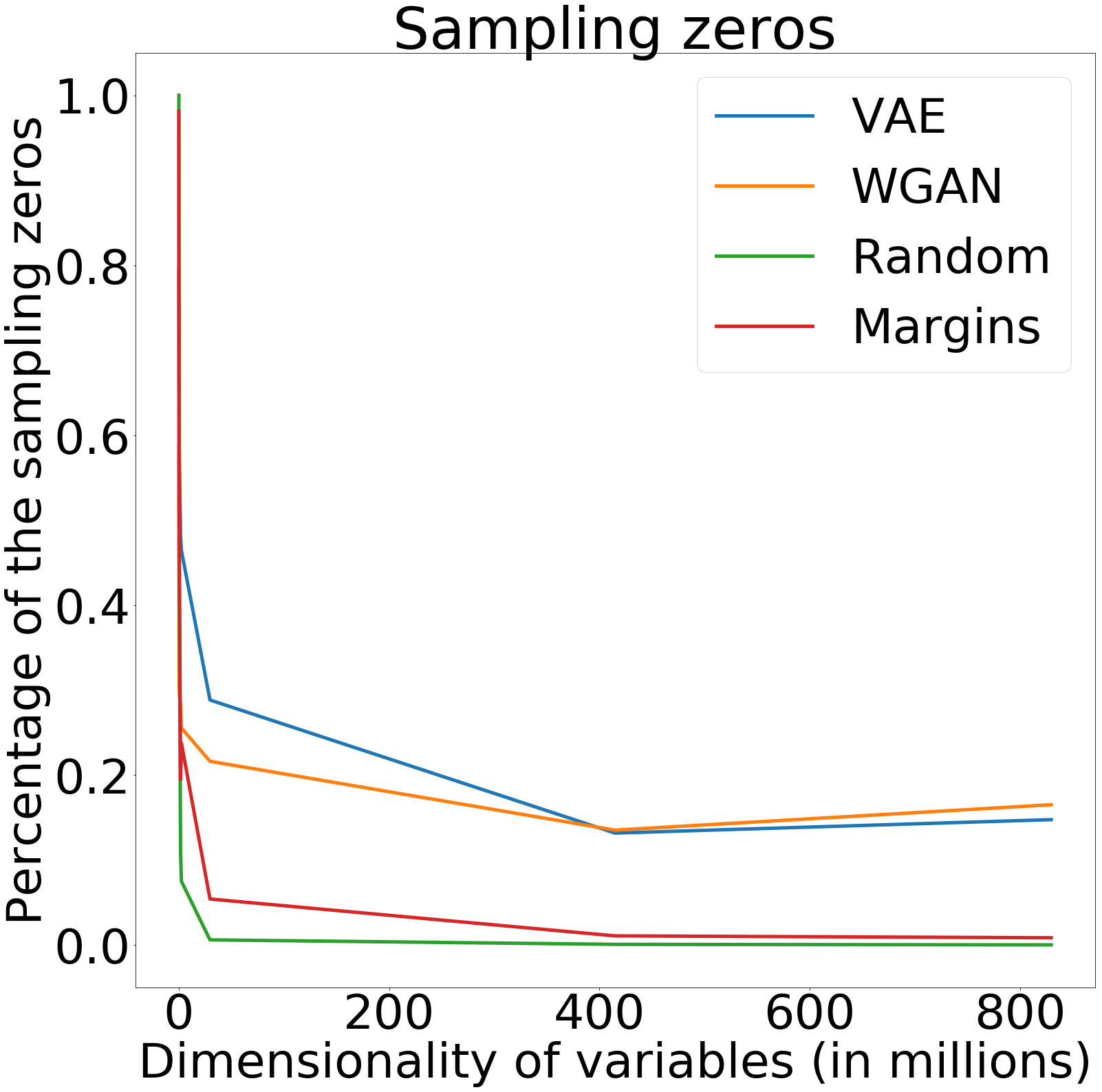}\hfill
        \includegraphics[width=.33\textwidth]{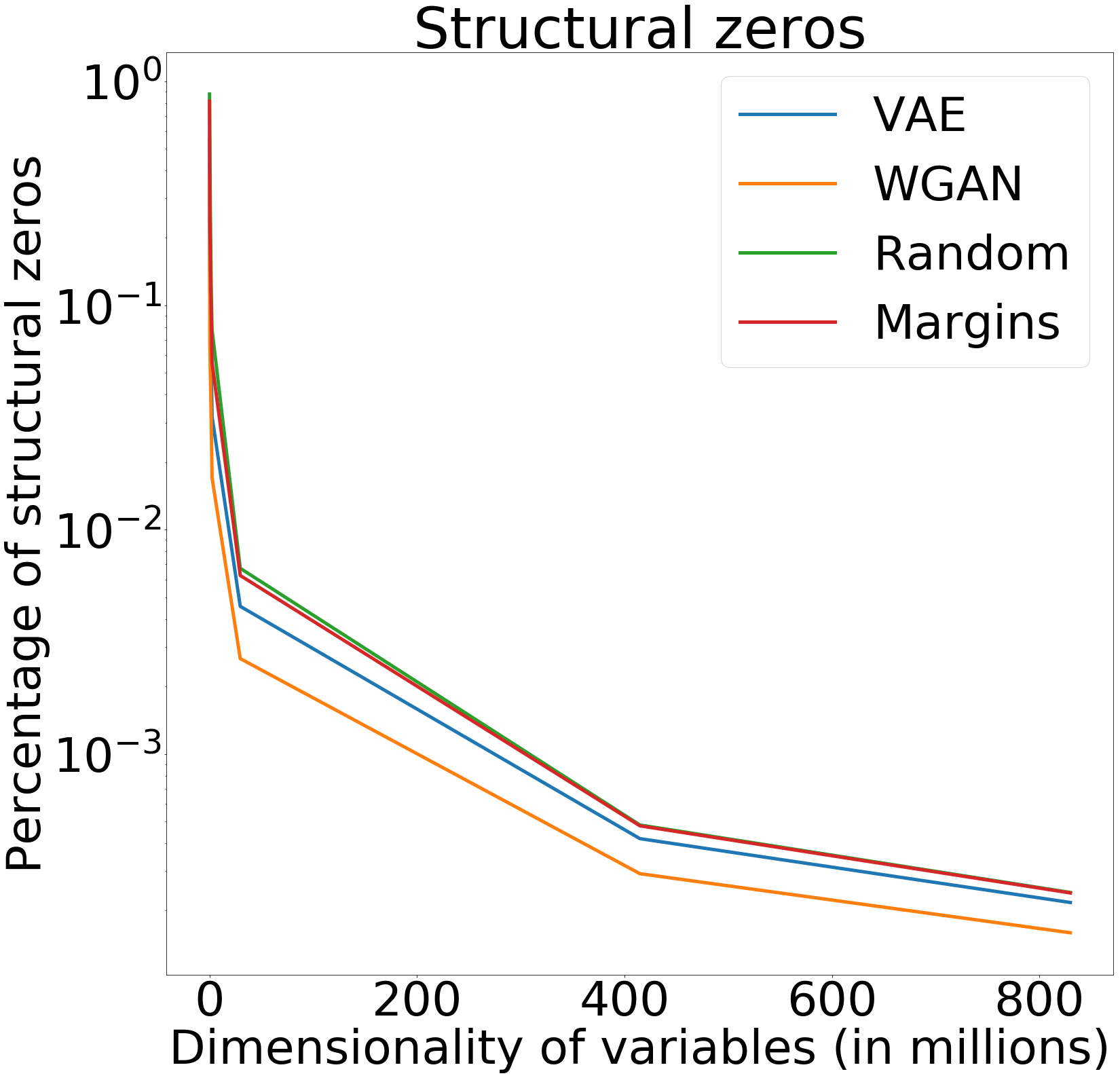}\hfill
        \includegraphics[width=.33\textwidth]{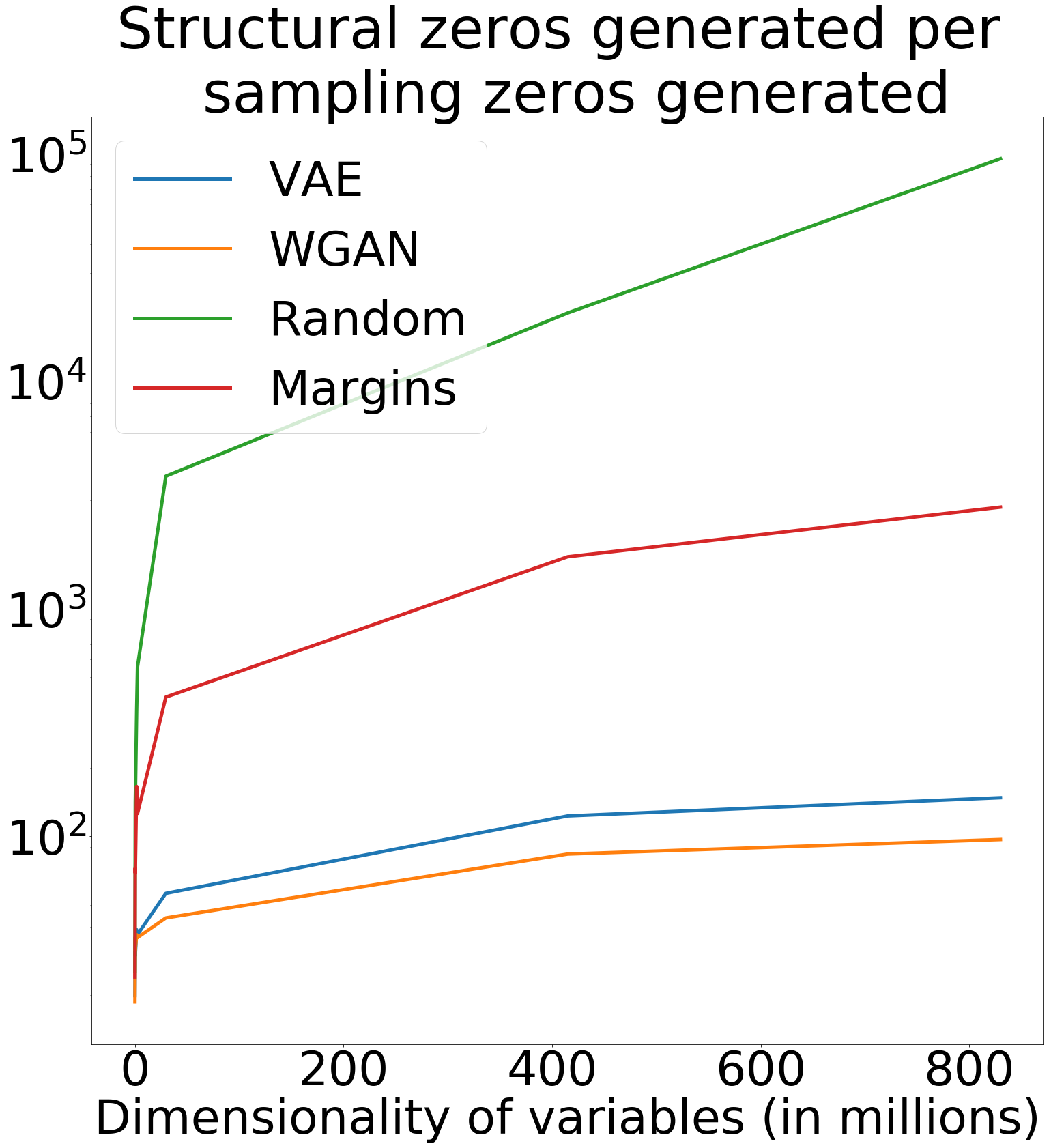}
    \end{subfigure}
    \caption{Dependence of sampling zeros, structural zeros and the ratio between them for different dimensions in variables with a fixed sample of 200k agents. Structural zeros and ratio in the logarithmic scale.}
    \label{fig:quality_vs_dim}
\end{figure}

Figure~\ref{fig:quality_vs_dim} shows another aspect of the same metrics, and we explore how these metrics change with the number of the dimensions of the underlying variables. In other words, how these metrics evolve as we include more variable or, potentially, more levels per variable. This is opposed to Figure~\ref{fig:sampling_structural_ratio} where the number of dimensions is fixed and we vary the number of simulated agents. We can see from the three different metrics that what is exposed on Figure~\ref{fig:sampling_structural_ratio} is replicated for any number of dimensions: The VAE and the WGAN are able to generate more sampling zeros, less structural zeros and the ratio is more favorable for the VAE and WGAN than the other models with the WGAN being the better model of the two.

In Table \ref{tab:ratio_percentages} we summarize some of the results of our models. The model performance is evaluated for different numbers of variable combinations. For the results in Table \ref{tab:ratio_percentages}, a sample of 200,000 individuals is drawn from the models and used as point of reference. The percentage of additional structural vs sampling zeros is considered by using the WGAN as a baseline although the ratio for the WGAN is also provided. It is clear that the WGAN is a superior model when using the proposed measure except in a single case where the VAE is approximately 5\% better than the WGAN. While it is not surprising that the marginal distribution or the random model have a poor ratio, it is surprising that the VAE is almost 50\% worse than the WGAN when investigated in a high dimensional setting.

\begin{table}[H]
    \small
    \centering
        \begin{tabular}{lllllllll}
        \toprule
        & \multicolumn{8}{c}{Number of variable combinations}                                     \\ \cmidrule{2-9} 
        & 9 801    & 274 428           & 352 836   & 1 646 568  & 2 469 852   & 29 638 224  & 414 935 136  & 829 870 272   \\
        Structural/Sampling zero &   &            &     &     &      &      &      &\\ 
        ratio for WGAN & 20.06   & 42.19            & 36.81    & 48.37    & 47.22     & 62.22     & 118.85     & 139.56     \\ \cmidrule{2-9} 
        & \multicolumn{8}{c}{Additional Structural/Sampling zero ratio using the WGAN as a base}                   \\ \cmidrule{2-9} 
        VAE                                     & 5.56\%  & \textbf{-4.59\%} & 6.50\%   & 10.93\%  & 10.04\%   & 27.48\%   & 40.99\%    & 44.74\%     \\
        Marginal distribution                   & 21.64\% & 92.91\%          & 118.68\% & 298.92\% & 225.47\%  & 633.36\%  & 1501.18\%  & 2217.19\%   \\
        Fully Random                            & 26.84\% & 146.15\%         & 281.10\% & 642.82\% & 1051.64\% & 5430.27\% & 23255.49\% & 170440.41\% \\
        \bottomrule
        \end{tabular}
    \caption{Percentage of the structural/sampling zero ratio using the WGAN as baseline. These figures are extracted from the sampling of 200,000 individuals.}
    \label{tab:ratio_percentages}
\end{table}

\subsection{Discussion}
Clearly, with reference to Table \ref{tab:ratio_percentages}, it is relevant to consider why the two models behave in this way. This type of behaviour has been widely studied and described in the machine learning literature \cite{theis_note_2016, sonderby_phd_2018} and can be attributed to the specification of the loss functions of these models. Particularly, the generative network of the VAE is trained to maximize the likelihood of the observed data. In practical terms this means that the inferred distribution will be \textit{mode covering}, i.e., the posterior distribution will try to cover all the modes of the distribution even when this implies assigning a higher probability to areas where this should not be the case. On the other hand, the training of GANs is associated with the minimization of the Jensen-Shannon (JS) divergence \citep{goodfellow_gan_2014}. This results in approximations that are \textit{mode seeking} and \textit{mode dropping} which, in practical terms, assign more probability than they should to specific modes of the true distribution. As explained in Section~\ref{sssec:GANs}, the choice of the Wasserstein-1 distance as the basis for the loss function, decreases the problem known as mode-collapse \citep{goodfellow_gan_2014}. Mode collapse occurs when several values of the noise variable are mapped to the same but plausible value by the generator and, as a consequence, still being accepted as a real sample by the discriminator. Much of the literature on deep generative models is concerned with alleviating the mode covering and seeking behaviour of these models.

\section{Summary and conclusions}
\label{sec:conclusion}
In the paper, we investigated the ability of two popular deep generative models---the Wasserstein Generative Adversarial Network (WGAN) and the Variational Autoencoder (VAE)---to recover sampling zeros from a high dimensional distribution. For this purpose, new metrics that relate the number of recovered sampling zeros with the number of generated structural zeros is proposed. The models are tested for a large-scale population synthesis application and it is found that, while the WGAN outperforms the other models in terms of prediction power when measured across several metrics, it however renders populations that are less diverse as compared to the VAE. 




The implication and impact of the presented research consist in the ability of the presented models to generate populations with a high degree of detail and to estimate the probability of rare combinations of features. These features are important for many agent-based model systems for which the ability to carry out long-term forecasts of detailed populations is essential.   

Several interesting new research directions have been identified during the writing of the paper. First, it will be interesting to apply even more complex architectures in neural networks and to combine deep generative models and alternative generative models. In particular is will be relevant to consider hierarchical deep generative models. Historically, hierarchical extensions of conventional population synthesis methods have proven useful in generating both households and individuals simultaneously \cite{casati_hierarchical_mcmc_2015, muller_2011_hipf} and in developing the deep generative models with hierarchical layers thereby representing a natural extension. Another research direction is to consider the capacity of the different models to project populations into the future. This will require the generated populations to be aligned with future targets possibly by stratified resampling. Another relevant research direction, which has emerged from the paper, is the modelling of attributes with an excessive number of categories. This represents a non-trivial challenge which is relevant for model systems where the populations are attached to a fine-grained geographical zone system. A hierarchical structuring of the zone system may provide a means of solving this problem, but more research is needed. Finally, the authors strongly believe that deep generative models can be used in a range of different contexts. As an example, they may be used to generate entire synthetic trip diaries although this will require modelling of logical constraints and intrinsic linkages between multiple activities and how these appear in space and time. 

\section*{Acknowledgements}

The research leading to these results has received funding from from the European Union's Horizon 2020 research and innovation programme under the Marie Sklodowska-Curie Individual Fellowship H2020-MSCA-IF-2016, ID number 745673. 


\nocite{tf}
\bibliographystyle{elsarticle-harv}
\bibliography{references.bib}

\end{document}